\documentclass[lettersize,journal]{IEEEtran}
\usepackage{amsmath,amsfonts}
\usepackage{algorithmic}
\usepackage{algorithm}
\usepackage{array}
\usepackage[caption=false,font=normalsize,labelfont=sf,textfont=sf]{subfig}
\usepackage{textcomp}
\usepackage{stfloats}
\usepackage{url}
\usepackage{verbatim}
\usepackage{graphicx}
\usepackage{cite}
\usepackage{xcolor}
\usepackage{hyperref}
\usepackage{listings}
\usepackage{blindtext}
\usepackage[T1]{fontenc}
\usepackage{float}
\usepackage{makecell}
\usepackage{diagbox}
\usepackage{color}
\usepackage{bbding}
\usepackage{multirow}
\usepackage{enumitem}
\usepackage{bbm}

\hypersetup{
	hidelinks
}

\lstdefinestyle{lfonts}{
	basicstyle   = \footnotesize\ttfamily,
	stringstyle  = \color{purple},
	keywordstyle = \color{blue!60!black}\bfseries,
	commentstyle = \color{olive}\scshape,
}
\lstdefinestyle{lnumbers}{
	numbers     = left,
	numberstyle = \tiny,
	numbersep   = 1em,
	firstnumber = 1,
	stepnumber  = 1,
}
\lstdefinestyle{llayout}{
	breaklines       = true,
	tabsize          = 2,
	columns          = flexible,
}
\lstdefinestyle{lgeometry}{
	xleftmargin      = 20pt,
	xrightmargin     = 0pt,
	frame            = tb,
	framesep         = \fboxsep,
	framexleftmargin = 20pt,
}
\lstdefinestyle{lgeneral}{
	style = lfonts,
	style = lnumbers,
	style = llayout,
	style = lgeometry,
}
\lstdefinestyle{python}{
	language = {Python},
	style    = lgeneral,
}

\begin{document}

\title{StarCraft+: Benchmarking Multi-agent Algorithms in \\Adversary Paradigm}

\author{Yadong~Li,
	Tong~Zhang,
	Bo~Huang,
	Zhen~Cui
	
	\IEEEcompsocitemizethanks{\IEEEcompsocthanksitem Y.Li, T.Zhang, B.Huang and Z.Cui are with School of Computer
		Science and Engineering, Nanjing University of Science and Technology, Nanjing, China.
		\protect\\
		E-mail: {liyadong, zhen.cui}@njust.edu.cn
		\IEEEcompsocthanksitem Y.Li are with School of Information Science and Engineering, Zaozhuang University, Zaozhuang, China.}
	\thanks{Manuscript received January xxx, 2024; revised xxx.xx, xxx.}
	\thanks{Corresponding author: Zhen~Cui.}
	\thanks{https://github.com/dooliu/SC2BA}
}

\markboth{Journal of \LaTeX\ Class Files,~Vol.~14, No.~8, Mar~2025}%
{Shell \MakeLowercase{\textit{et al.}}: A Sample Article Using IEEEtran.cls for IEEE Journals}

\IEEEpubid{0000--0000/00\$00.00~\copyright~2021 IEEE}

\maketitle

\begin{abstract}
Deep multi-agent reinforcement learning (MARL) algorithms are booming in the field of collaborative intelligence, and StarCraft multi-agent challenge (SMAC) is widely-used as the benchmark therein. However, imaginary opponents of MARL algorithms are practically configured and controlled in a fixed built-in AI mode, which causes less diversity and versatility in algorithm evaluation. To address this issue, in this work, we establish a multi-agent algorithm-vs-algorithm environment, named StarCraft II battle arena (SC2BA), to refresh the benchmarking of MARL algorithms in an adversary paradigm. Taking StarCraft as infrastructure, the SC2BA environment is specifically created for inter-algorithm adversary with the consideration of fairness, usability and customizability, and meantime an adversarial PyMARL (APyMARL) library is developed with easy-to-use interfaces/modules. Grounding in SC2BA, we benchmark those classic MARL algorithms in two types of adversarial modes: dual-algorithm paired adversary and multi-algorithm mixed adversary, where the former conducts the adversary of pairwise algorithms while the latter focuses on the adversary to multiple behaviors from a group of algorithms. The extensive benchmark experiments exhibit some thought-provoking observations/problems in the effectivity, sensibility and scalability of these completed algorithms. The SC2BA environment as well as reproduced experiments are released in \href{https://anonymous.4open.science/r/SC2BA-E853}{Github}, and we believe that this work could mark a new step for the MARL field in the coming years.
\end{abstract}


\section{Introduction}
\label{sec:introduction}

\IEEEPARstart{D}{eep} reinforcement learning (RL) has shown great potential to solve arbitrary sequential decision-making problems and has been successfully applied in many domains, such as traffic control~\cite{shang2025graph, wang2024multi, zhang2019cityflow}, electric game~\cite{mnih2015human, vinyals2019grandmaster}, resources allocation~\cite{zheng2024multi, wu2024adaptive}, robot path planning \cite{wang2020mobile, chung2024learning}, image classification~\cite{mousavi2019multi} and stock market \cite{zou2024novel, huang2024multi}. Training RL agents requires a substantial amount of data, which consists of abundant action sequences and rewards, while acquiring such data in the real world can be time-consuming and costly. Hereby, the simulated multi-agent systems (MAS)~\cite{stone2000mas} are urgently required to provide sufficient synthetic data in different application scenarios for reinforcement learning research.

Various research efforts have been devoted to the development of effective MAS, where simulating multi-agent interaction among agents plays a crucial role. Specifically, the representative interaction relationships between agents are cooperation and competition in MAS. In the early works, multi-agent simulation environments mainly focus on cooperative tasks, such as Bi-DexHands~\cite{chen2022towards}, Multi-Agents Mojoco~\cite{peng2021facmac}, RWARE~\cite{papoudakis2020benchmarking}, PressurePlate~\cite{McInroe2022pressureplate}, and Flatland~\cite{mohanty2020flatland}. These simulation environments are specifically designed to address real-world production tasks, such as two-arm collaboration, robot movement control, and cargo loading. Thus, they greatly propel the application of robotics in this domain. More simulation environments encompass both cooperation and competition, such as LBF~\cite{papoudakis2020LBF}, MALMO~\cite{johnson2016malmo}, MPE~\cite{lowe2017multi}, MAgent~\cite{zheng2018magent}, DM Lab~\cite{beattie2016deepmind}, PommerMan~\cite{resnick2018pommerman}, Google Research Football~\cite{kurach2020grf}, Go Bigger~\cite{zhang2023gobigger}, MA Particle environment~\cite{mordatch2018mpe}, Neural MMO~\cite{suarez2021neural}, and SMAC~\cite{samvelyan2019smac}. Within these environments, apart from collaborative tasks, adversarial settings are also integrated. Agents must acquire the skills to comprehend cooperation within their camp and competition between camps to accomplish targeted tasks and attain higher rewards. This task augmentation increases the complexity of multi-agent environments, enabling intricate modeling of dynamic game scenarios and fostering advancements in multi-agent research.

Although considerable success has been achieved by existing MAS, a prominent issue arises in most benchmark simulation environments that opponent units are designed to be controlled by built-in AI bots. These pre-set rules of AI bots may result in insufficient diversity for multi-agent training. Commonly, the agents may be trained with biases toward exploiting certain weaknesses of the pre-set strategies of opponents. However, when facing other opponents using different combat rules, these trained agents usually obtain poor performance. Moreover, due to the monotony of the enemy's strategy, it may greatly limit the strategy diversity of MARL models if most policy explorations do not yield additional combat benefits. 
Hence, to construct an MAS that better supports the evaluation/learning on MARL models, the following issues should be well addressed.

\begin{itemize}[leftmargin=*]
	\item[-] \textbf{Opponent non-monotonicity}. As the design of built-in AI bots is primarily influenced by the subjective cognition of designers, some latent fatal rules/limitations may be exploited during agent training, which would cause the biases of learned policies. Further, due to monotonic opponent, trained agents tends to fall in a limited behavior distribution, not well catering to scene diversity~\cite{lanctot2017unified}.
	\IEEEpubidadjcol
	\item[-] \textbf{Opponent evolvability}. For static strategies/rules of opponents, the agents often quickly acquire winning techniques, even for well-designed built-in AI bots. It means that the agents might learn an optima of simplex opponent, but difficult to generalize to more diverse cases. An essential reason should be the lack of opponent dynamic variations. In fact, for many real-world adversarial environments, the opponents have the evolutive ability during adversary process, which is just ignored by most MAS. 
	\item[-] \textbf{Adversarial fairness}. When both game parties (i.e., multi-agent teams) are evolvable, their adversarial fairness evaluation becomes a key issue. There exist many factors, such as initial states, model sizes, parameter settings and force layouts, that affect the final game results. Hence, to ensure the evaluation as fair as possible, it is important to bypass the influence of those external factors and reveal the inherent ability of multi-agent algorithms.
	\item[-] \textbf{System Usability}. The MAS system should be user-friendly and easy to operate, offer convenient invocation and support researchers in customizing scenarios. It also allows researchers to easily implant the proposed algorithms for comparison and analysis.
\end{itemize} 

To address these issues aforementioned, we build an algorithm-vs-algorithm environment, named StarCraft II Battle Arena (SC2BA), as a benchmark to facilitate the MARL research. To enrich the diversity of environment, we attempt to endow enemy units with more abundant behaviors. To encourage agents to explore more complex winning strategies, we enhance dynamic adversary to the opponent. More straightforwardly, the opponent (a.k.a blue team) is no longer controlled by fixed built-in AI bots but rather various MARL algorithms which could boost the adversarial ability of blue team. To this end, we configure two types of game modes: dual-algorithm paired adversary mode and multi-algorithm mixed adversary mode. The former conducts the adversary of pairwise algorithms while the latter focuses on the adversary to multiple behaviors from a group of algorithms. Both of them are used to better support the evaluation of the generalization, diversity and adversarial ability of multi-agent learning algorithms. Further, by using these modes, the MARL models (w.r.t not only red team but blue team) can be trained to possess better adaptability to unseen opponents in contrast to the constant built-in AI opponent.
In order to evaluate multi-agent algorithms on scenario changes, we also provide a series of meticulously designed combat scenarios, incorporating both symmetric and asymmetric setups. These scenarios pose challenges for algorithms dealing with high dimensional inputs, partial observability, rich dynamics, and the coordination of independent agents in joint battles. 

Building upon the popular real-time strategy (RTS) game StarCraft II, SC2BA aims to provide a comprehensive evaluation platform for MARL algorithms and foster advancements in the field of multi-agent reinforcement learning. By providing a robust and scalable battle platform, along with algorithm-vs-algorithm adversary modes, diverse combat scenarios and evaluation metrics, SC2BA aims to facilitate the exploration and improvement of MARL algorithms. It is our hope that SC2BA will serve as a catalyst for further advancements in the field, fostering innovation and pushing the boundaries of Deep MARL research. To summarize, the contribution of this work is four folds: 
\begin{itemize}[leftmargin=*]
	\item[\romannumeral1)] We develop the SC2BA which is a convenient, scalable, and unrestricted combat platform. Unlike the previous environments where the opponents are controlled by embedded AI bots, this environment allows both multi-agent teams to be controlled by designed MARL algorithms in a continuous adversarial paradigm. Compared to the built-in AI mode, SC2BA enables a more comprehensive evaluation of algorithm performance and facilitates the training of agents with diverse and complex strategies.
	\item[\romannumeral2)] We introduce diverse adversary modes and meticulously design a series of challenging battle scenarios. These modes enhance diversity, adversary and complexity of the environment, encouraging agents to explore more robust policies. Moreover, these not only enrich the evaluation ways but also provide the verification on the generalization, diversity and adversarial ability for different multi-agent algorithms.
	\item[\romannumeral3)] We develop and open-source APyMARL framework with easy-to-use interfaces/configurations and flexible modules, so as to serve SC2BA. APyMARL is modular, scalable, built on PyTorch, and serves as a research platform for studying some interesting challenges of MARL.
	\item[\romannumeral4)] We conduct extensive experiments by dissecting every adversary mode as well as comparing those advanced methods. The experiments validate the effectiveness of our proposed adversary modes and meantime report results of several cutting-edge algorithms, serving as benchmarks for evaluation. More importantly, by running algorithm-vs-algorithm adversary modes, we obtain some extraordinary observations on the existing multi-agent algorithms as well as battle scenarios, which could benefit future MARL research.
\end{itemize} 

\section{Related Work}\label{sec:relatedwork}

MARL simulation environments are typically classified into cooperative, competitive, and mixing settings based on specific requirements~\cite{zhang2021multi}. Cooperative MARL constitutes a great portion of MARL environments, where all agents collaborate with each other to achieve some shared goal. MA cooperative tasks primarily revolve around domains such as robot team navigation~\cite{yang2018keeping, shamsoshoara2019distributed, tovzivcka2019application}, smart grid operation~\cite{dall2013distributed}, control of mobile sensor networks~\cite{cortes2004coverage}, and other related areas. These tasks are specifically designed to address real-world application problems. Therefore, many scholars designed excellent simulation environments for MARL cooperative research, making profound contributions to the advancement of the MARL domain. To advance the capabilities of robots in achieving human-level sophistication of hand dexterity and bimanual coordination, Yang et al. proposed Bi-DexHands~\cite{chen2022towards}. It offers a series of intricate tasks involving dexterous hands control, which demand agents to acquire refined operational skills. To advance research in the field of cooperative control of unmanned devices, researchers have introduced several exceptional simulation environments.  AirSim~\cite{shah2018airsim} is an open-source simulation platform developed by Microsoft for research and development in autonomous systems. It provides a realistic virtual environment that enables researchers and developers to test, validate, and refine their algorithms and models before deploying them on real-world autonomous vehicles or robots. Additionally, Pedra~\cite{anwar2020autonomous} is a programmable engine for reinforcement learning in the domain of unmanned aerial vehicles~(UAVs). It constructs a collection of indoor and outdoor 3D simulation scenarios for studying UAV control and decision-making problems. 

Competitive MARL settings are typically modeled as zero-sum games. MA competitive tasks primarily focus on classic games, including board games and poker games. At present, there are two notable competitive MARL environments: Go and Texas Hold'em Poker, which are archetypal instances of multiplayer perfect-information and partial-information extensive-form games, respectively. Go~\cite{silver2017mastering} is an ancient and intellectually captivating strategy board game, often regarded as one of the oldest, most complex, and highly challenging games in existence. Texas Hold'em Poker~\cite{bowling2015poker} usually entails the participation of two or more players, with each player initially receiving two face-down private cards. This is followed by three rounds of community cards being placed face-up on the table, resulting in a total of five shared cards. Agents participating in this game must make strategic decisions, considering the limited information available, and choose from possible actions: check, call, raise, and fold.

In comparison to the former two settings, mixed-setting environments are more challenging. After achieving remarkable advancements in RL within board game environments, researchers turned their attention to electronic games and have developed excellent MARL simulation environments, such as Atari~\cite{mnih2013atari}, Super Mario~\cite{kuhn2014mario}, Gran Turismo racing games~\cite{wurman2022gran}, Doom~\cite{Kempka2016ViZDoom}, GRF~\cite{kurach2020grf}, Dota2~\cite{berner2019dota}, Honor of Kings~\cite{ye2020mastering}, StarCraft~\cite{vinyals2019grandmaster}. Among them, Real-Time Strategy (RTS) games have garnered considerable attention. After AlphaZero~\cite{silver2017mastering} emerged as the strongest general-purpose AI in chess, DeepMind shifted its focus to the RTS game StarCraft II. Vinyals et al.~\cite{vinyals2017sc2le} introduced the StarCraft II Learning Environment~(SC2LE), which focuses on tackling the full game of StarCraft II by centralized controlling hundreds of agents to defeat competitors. Shortly thereafter, Vinyals et al.~\cite{vinyals2019grandmaster} trained AlphaStar using SC2LE, showcasing master-level proficiency and achieving the milestone of defeating professional human players for the first time. However, SC2LE focused on the 1v1 mode, where agents act the role of player and have centralized control over all combat units. This setup is fundamentally unrealistic in the real world. Therefore, Samvelyan et al.~\cite{samvelyan2019smac} proposed SMAC based on SC2LE, focusing on decentralized micromanagement challenges instead of centralized control over the entire game. The design of the SMAC environment simplifies the validation of MARL algorithms and reduces the demand for high computational resources. This accessibility has encouraged greater participation from researchers, significantly fostering the development of MARL research.

However, in the aforementioned works, the enemy units were controlled by the built-in AI bots and the agents engaged in training and testing against adversaries with identical behavioral patterns. This setting prevents the enemy forces from exerting pressure on the agents, keeping them in a constant disadvantaged position. Instead, it allows the agents to easily find ways to defeat their opponents and achieve victory. Taking SMAC as an example, Ellis et al.~\cite{ellis2022smacv2} discovered that retained only agent ID and timestep information in the observations, QMIX~\cite{rashid2020qmix} and MAPPO~\cite{yu2022mappo} algorithms achieved performance comparable to using all available information. This is highly disadvantageous for agents to learn robust strategies. Contrarily, Emergent tools~\cite{baker2019emergent} demonstrate the potential for simultaneous online learning by both factions. The hiders and seekers exhibit extraordinary strategy and counter-strategy in their game against each other. Inspired by this, we develop a new adversarial environment named SC2BA. Different from previous works, our work propose an online game platform with two competing teams, where agents from both sides could optimize themselves to struggle for victory. Accordingly, the algorithm-vs-algorithm adversary could be run directly and their performance could be validated more comprehensively.

\section{SC2BA System Overview}
\label{sce:sc2ba_overview}

\begin{figure}[h]
	\centering
	\subfloat{
		\includegraphics[width=1.0\linewidth]{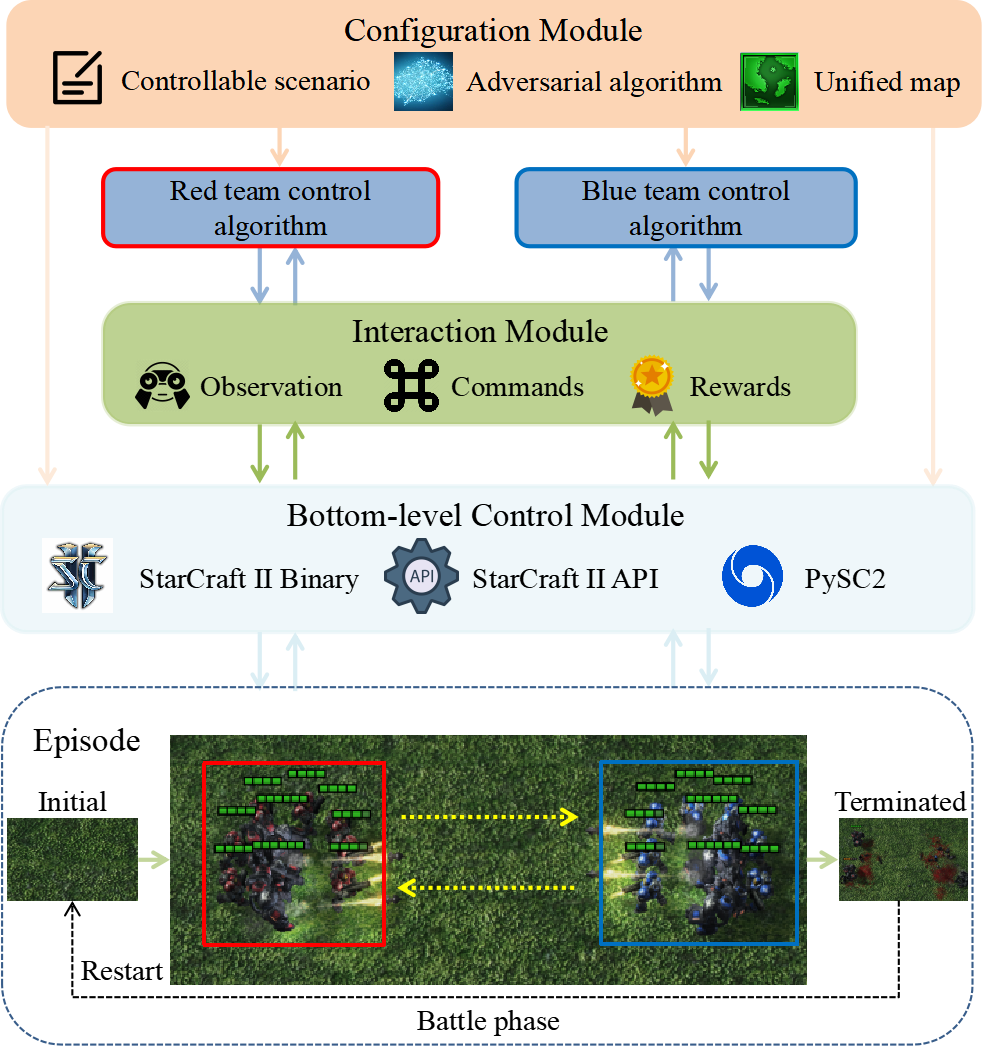}
	}
	\caption{The details of SC2BA platform, shown with its sub-module: configuration module, interaction module, bottom-level control module. Please see details in Section~\ref{sce:sc2ba_overview}}
	\label{sc2ba_framework}
\end{figure}

The SC2BA system schema, as shown in Fig.~\ref{sc2ba_framework}, is designed with modularity to facilitate the adversary between algorithm and algorithm. It redefines the configuration of battle scenarios and interactions between agents and environment. In SC2BA, users can easily customize the desired battle scenarios through the configuration module. In each episode, the agents of red and blue teams are individually controlled by two separate models driven by MARL algorithms. Through the interaction module, both teams can access their respective observation information in the environment. Tactical actions on both teams are made through their employed algorithms, according to their observation information. The action commands will be fed back to agents and executed for the next step. During each battle episode, the control algorithms of both teams only interact with the interaction module, allowing the system to concentrate on battle policies and further reveal the capabilities of multi-agent algorithms. In the following subsections, we will elaborate on the details of each module.

\subsection{Configuration Module}
\label{subsec:configuration}

The configuration module redefines the creation of scenarios based on SMAC, allowing users to adjust the scene in a user-friendly manner. The settings of interactive algorithms, battle force, multi-agent attributes as well as scene elements are completely formatted with text prompts, thus the operability of scene configuration could be greatly enhanced and the burden of algorithm developers is reduced.

\textbf{Unified map file}. We integrate all combat units into one map, allowing any battle scenarios to be implemented within this unified map, while the previous paradigm employs an individual map file for each scene, which makes scene definition/modification tedious and error-prone. 

\textbf{Controllable scenario}. To facilitate the setup of various aspects in a battle scene, our system allows for effortless creation and modification of these scenes. The system supports adjustable parameters for battle scenarios, including the count of agents, team composition, start position, movable areas, and other relevant factors.

\textbf{Adversarial Algorithms}. We can assign distinct multi-agent algorithm models to control the agents on both teams using commands. The starting point of algorithm models is set to random sampling or individual well-trained models (e.g., trained with built-in AI bots). During the training stage, we can set both teams to be controlled by algorithms capable of continuous evolution for injecting dynamic variations. Additionally, we can configure multiple well-trained models to alternately control the enemy troops, which could enrich the behavioral diversity to some extent.

\subsection{Interaction Module}
\label{subsec:interaction}

The interaction module is responsible for handling all interaction operations between both control algorithms and the environment, including transmitting observations, states, action instructions and reward values. Additionally, SC2BA follows the design principles of gym~\cite{brockman2016gym} and develops easy-to-use operation settings to simplify the interaction. The interaction mainly lies in three-fold:

\textbf{Observation exchange}. For a more convenient exchange, the observations/states of agents are structured in a vectorized format. Hereby, the interaction module only needs to transmit those vector signals. Accordingly, multi-agent algorithms can easily access structured signals, which makes MARL researchers focus on solving cooperative and adversarial tasks.

\textbf{Action interaction}. For the action interaction between algorithms and environment, we simplify the vast and complex instruction set in StarCraft II, convert these discrete actions into executable instructions, and then pass them to the game control module.

\textbf{Rewards feedback}. The interaction module receives the reward information from the control terminate based on the current battle situation, and then feeds back to multi-agent algorithms. In the adversarial mode, the feedback consists of the evaluation scores of both parties, i.e., damaging/massacring enemies as well as penalties for self-inflicted harm and death.

\subsection{Bottom-Level Control Module}

The bottom-level control module is responsible for managing the dynamic evolution of the system and offers an API that the interaction module can leverage. Through instructions, the interaction module can feed action commands to the game engine and access information from the environment, including the positions, health, shield status, and other relevant data of the agents. In addition, it has the capability to generate appropriate battle scenarios based on configuration information while maintaining complete transparency with respect to the team control algorithms.

\textbf{Linux SC2 binary}\footnote{StarCraft II version: SC2.4.6.2.6923}. The Linux version binary package of SC2 provides a game engine specifically designed for Linux operating systems. This game engine empowers us to seamlessly implement all our ideas. Furthermore, the Linux version engine does not require rendering graphics during gameplay, greatly improving sampling speed.

\textbf{SC2 API}\footnote{SC2API: https://github.com/Blizzard/s2client-proto}. The SC2 API is an interface that offers complete external control of SC2 game. It is designed to connect with the above interaction module to exchange information that algorithms require. 

\textbf{PySC2}\footnote{PySC2: https://github.com/google-deepmind/pysc2}. PySC2 is a Python toolkit that provides a wrapper for the SC2 API, enabling direct utilization of SC2 interface through Python code. This simplifies the development of reinforcement learning environments based on SC2, as it allows developers to concentrate on how to learn algorithmic strategies without expending additional effort on studying game engine manuals.

\textbf{Procedural control}. To streamline the training/testing process, we automatically control the game stages through some code commands. This allows us to easily generate battle episodes with diverse unit compositions and environmental conditions based on specific configuration information. Furthermore, it automatically restarts a new episode upon the current one is terminated.

\section{The SC2BA Environment}

SC2BA provides an algorithm-vs-algorithm adversary evaluation platform by offering a powerful and scalable battle environment, along with two adversary modes, various combat scenarios, and other designs. Below, we introduce the details of each part of SC2BA.

\subsection{Adversary Modes}

To better verify the diversified and adversarial ability of MARL algorithms, we design two adversary modes: dual-algorithm paired adversary mode and multi-algorithm mixed adversary mode.

\textbf{Dual-algorithm paired adversary mode}. It introduces algorithm-vs-algorithm combat to enhance dynamic adversarial interactions, thus encouraging multi-agent to explore more robust winning strategies. In each scenario, agents are divided into two teams, each of which is controlled by a MARL algorithm. The two teams can continuously learn action policies and adapt to dynamic changes of the opponent. This requires MARL algorithms to possess robust adaptive and learning capabilities to gain an upper hand against ever-evolving opponents. Thus, this mode can evaluate the adversarial capability between two MARL algorithms.

\textbf{Multi-algorithm mixed adversary mode}. This mode is used to further evaluate the ability of a multi-agent algorithm when confronted with more diverse opponent actions. To implement this, we integrate multiple well-trained MARL models into SC2BA to alternately control the enemy troops. The opponent model is randomly selected from specified MARL algorithms for each episode, making the opponent's actions completely transparent and unpredictable. Thus, the agent algorithm should adapt to different opponents, which boosts them to master a wider range of tactical approaches. 

\begin{table*}[!htb]
	\setlength{\abovecaptionskip}{-0.15cm}
	\setlength{\belowcaptionskip}{-0.2cm}
	\renewcommand{\arraystretch}{1.3}
	\caption{The current SC2BA scenarios. The system may be extended to other scenarios. }
	\label{SC2BA_scenarios}
	\centering
	\begin{tabular}{c c c}
		\hline
		Name & Red Units & Blue Units \\
		\hline
		3m & 3 Marines & 3 Marines \\
		8m & 8 Marines & 8 Marines \\
		25m & 25 Marines & 25 Marines \\
		MMM & 1 Medivac, 2 Marauders \& 7 Marines & 1 Medivac, 2 Marauders \& 7 Marines \\
		\hline
		2s3z & 2 Stalkers \& 3 Zealots & 2 Stalkers \& 3 Zealots \\
		3s5z & 3 Stalkers \& 5 Zealots & 3 Stalkers \& 5 Zealots \\
		1c3s5z & 1 Colossus, 3 Stalkers \& 5 Zealots  & 1 Colossus, 3 Stalkers \& 5 Zealots\\
		\hline
		5m\_vs\_6m & 5 Marines & 6 Marines \\
		10m\_vs\_11m & 10 Marines & 11 Marines \\
		MMM2 & 1 Medivac, 2 Marauders \& 7 Marines & 1 Medivac, 3 Marauders \& 8 Marines \\
		\hline
	\end{tabular} 
\end{table*}

\begin{table*}[!htb]
	\setlength{\abovecaptionskip}{-0.15cm}
	\setlength{\belowcaptionskip}{-0.2cm}
	\renewcommand{\arraystretch}{1.3}
	\caption{The agent parameters used in SC2BA .}
	\label{SC2_units}
	\centering
	\begin{tabular}{c|c|c|c|c|c|c|c|c}
		\hline
		Name & health & shield & Race & Attack range & Attack damage & Attack speed & Move speed & Armor \\
		\hline
		Marine & 45 & - & T & 6 & 6 & 0.86 & 2.25 & Light \\
		Marauder & 125 & - & T & 6 & \thead{10 | vs Armored 20} & 1.5 & 2.25 & Armored \\
		Medivac & 150 & - & T & 6 & - & - & - & Armored \\
		\hline
		Zealot & 100 & 50 & P & 6 & 16 & 1.2 & 2.25 & Light \\
		Stalker & 80 & 80 & P & 6 & \thead{13 | vs Armored 18} & 1.87 & 2.25 & Armored \\
		Colossus & 200 & 150 & P & 6 & \thead{20 | vs Light 30} & 1.5 & 2.25 & Armored \\
		\hline
	\end{tabular} 
\end{table*}

\subsection{Scenario}
\label{smab_scenatios}

In a regular full game of SC2, the players of two camps compete for resources, construct buildings, train armies, and eliminate all enemy buildings to achieve ultimate victory. Differ from the full game of SC2, we design a flexible team-adversarial game with the following specifications: each player controls a single combat unit instead of managing a complex macro-level strategy with an entire army; each team consists of multiple units that collaborate to fight against the opposing team to achieve final victory.

To evaluate MA algorithms on scenario changes, we provide a series of meticulously designed combat scenarios, incorporating symmetric and asymmetric setups. The currently defined scenarios are listed in Table~\ref{SC2BA_scenarios}. They can be used to evaluate the ability of MARL algorithms when suffering diverse behaviors, rich dynamics, unequal troops and difficult collaboration. 

\textbf{Symmetric scenarios}. In symmetric scenarios, our goal is to create a balanced competitive platform. The composition and quantity of units for both teams are equal, ensuring a relatively balanced starting point. Symmetric scenarios are devoted to revealing MARL algorithms' inherent ability in learning coordination strategy and execution skills. 

\textbf{Asymmetric scenarios}. Asymmetric scenarios are more challenging due to non-equilibrium forces/layouts. For example, the enemy team possesses a quantity advantage in terms of troop strength, thereby intensifying the challenge of achieving victory. This setup necessitates MARL algorithms to demonstrate superior-level cooperation and remarkable operation skills to overcome asymmetric challenges. 

\subsection{State and Observation}

In each time step of battle episode, agents can obtain observation information within their perception field from the environment. As used in SMAC, the sight range of all agents is also set to 9. It means that agents are partially observable in this setting. According to the sight range, we have the observation attributes of each agent as follows:

\begin{lstlisting}[style = python]
	observation_data = {
		'movements': [north, south, east, west],
		'enemies': [[enemy_id, distance, relative_x, relative_y, health, shield, unit_type], ...],
		'allies': [[distance, relative_x, relative_y, health, shield, unit_type], ...],
		'personal': [health, shield, unit_type] 
	}
\end{lstlisting}

The adversary states record the information of all agents from both teams at one-time slice, including the position of all units relative to the central point and other relevant features of agents. Please note that the adversary states are only available during the training phase. The details of the adversary state are outlined below.

\begin{lstlisting}[style = python]
	state_data = {
		'enemies': [[health, weapon_cd, relative_x, relative_y, shield, unit_type], ...],
		'allies': [[health, relative_x, relative_y, shield, unit_type], ...],
	}
\end{lstlisting}

\subsection{Action}
To simplify the vast and complex instruction set in SC2, we utilize a discrete set of actions that are automatically converted into executable instructions within the game engine. The action space is divided into three categories: move, attack, stop/no-op. The move action further contains four directions: north, south, east and west. The living agents can perform attack actions (when hostile agents are viewed and within attack range), or moving operations, or even stop operation (when choosing to maintain their current state). Dead agents can only perform no-op action.

In the simulated environment, the attack ranges for all units are set to a fixed value of 6, same to SMAC. If an agent executes the attack action out of actual attack range, it can not directly open fire. Instead, it triggers a built-in attack-move macro-actions on the target unit. When the enemies reach the attack radius, The agent will automatically go close to the specified enemy and start firing. 

Battle units possess a diverse range of attributes and exhibit mutual counters. First, units possess different armor types: light and armored(heavy armor). Second, the arsenal encompasses a variety of weapons, with certain weapons bearing special effects. For instance, the marauder's punisher grenades inflict double damage against units with heavy armor, while the colossus is equipped with dual thermal lances that inflict area damage and extra damage against light-armored units. 

\subsection{Reward}

To promote adversarial dynamics among algorithms, we redefine the reward function within SMAC. The reward function comprises two primary elements: immediate rewards and cumulative rewards. Firstly, for immediate rewards, we adhere to the design principles established in SMAC, where agents receive response rewards based on the current battle situation in each time step. These rewards consist of damaging/killing enemies as well as penalties for self-inflicted harm and death. Furthermore, with regard to the episodic rewards, we introduce penalties for failures and draws. For example, if neither of teams manages to eliminate all enemies within the pre-specified time limit, both teams receive a penalty reward. This setting carries meaning as it inspires an active participation mindset in both teams, encouraging them to intensely pursue victory rather than avoiding conflict out of fear of failure.

\subsection{Layout}

In SC2BA, our primary goal is to construct a fair competition platform that allows two distinct algorithms to effectively demonstrate their adversarial capabilities. When engaging in battles between these paired MA algorithms, it is vital to maintain consistent and reliable results by employing identical layouts. To create a reliable assessment platform, we uphold several guiding principles, such as matching combat forces, symmetrical spatial stations and identical observation views.

\textbf{Matched battle forces}. It is widely recognized that achieving an equal layout primarily requires ensuring equivalent or nearly equal combat forces for both teams. The discrepancy in combat forces significantly influences the outcome of battles. For example, in SC2, it is impossible to gain victory when nine marines combat with ten marines. Furthermore, the side with fewer soldiers can only eliminate a limited number of opposing units, rather than an equivalent number of nine.

\textbf{Symmetric space stations}. In battles, the positioning of each individual agent within a team significantly influences the result and has a direct impact on subsequent formations and combats. For instance, based on an analysis of massive battle videos, we have discovered that a vertical arrangement of three soldiers, all facing the opponent, is more likely to secure victory compared to a horizontal arrangement. Therefore, we create multi agents of both teams in a central symmetric fashion, guaranteeing equality in the positions and formations.

\textbf{Identical observation views}. To mitigate the impact of team positioning on observational perspectives and subsequent decision-making processes of the models, we adopt an identical observation view for both teams. Drawing inspiration from games such as "Honor of Kings" (also known as "King of Glory"), where a mirrored map is employed, the red and blue teams share identical perspectives. This design choice eliminates positional bias, allowing players on both sides to operate under fair and unbiased conditions.

\section{Adversarial PyMARL}\label{sec:lib}

\begin{figure*}[!htb]
	\centering
	\subfloat{
		\includegraphics[width=1.0\linewidth]{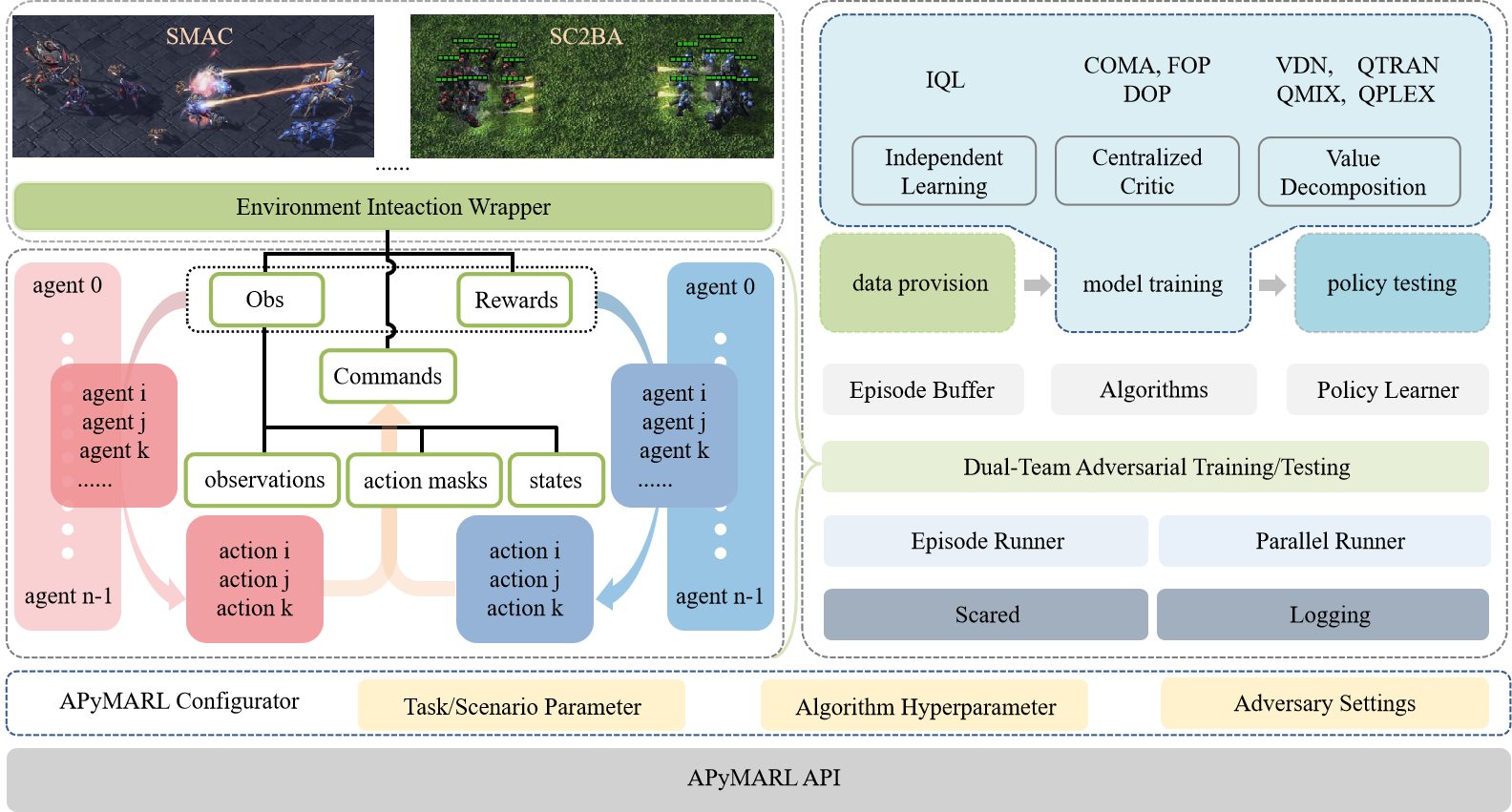}
	}
	\caption{Overview of APyMARL framework. More details can be found in Section \ref{sec:lib}.} 
	\label{APyMARL_framework}
\end{figure*}

Currently, most MARL learning frameworks~\cite{hu2021pymarl2, raffin2021stable, hoffman2020acme, pretorius2021mava, hu2023marllib} focus on single-team multi-agent training, i.e., utilizing built-in AI bots to control opponent units. For instance, PyMARL~\cite{rashid2020qmix}, the most popular MARL algorithm library, only supports the SMAC environment and trains multi-agent by fighting against the built-in AI bots. Extended from the PyMARL framework, the EPyMARL~\cite{papoudakis2020epymarl} supports four different learning environments, but it still focuses solely on single-team multi-agent collaboration tasks and cannot manipulate dual-team adversarial agents. This setting limits their ability to support the paradigm of dual-team multi-agent adversarial learning. 
Accordingly, these learning frameworks do not provide those interfaces for algorithm-to-algorithm adversarial training/testing. As a result, the limitation hinders the development of dual-team multi-agent adversary research and impedes comprehensive assessments of MARL algorithms. 

In this regard, we attempt to develop an algorithm-to-algorithm adversarial framework, called Adversarial PyMARL (APyMARL), in order to support dual-team agents combat tasks, disentangle the environment and the algorithm, and provide a standardized testing suite for MARL. APyMARL incorporates several new features, while leveraging the key advantages of PyMARL. These new features mainly include a standardized multi-agent environment interaction wrapper, dual-team multi-agent adversarial training/testing controllers, as well as a flexible configurator.

\textbf{Multi-agent environment interaction wrapper}. A standardized agent-environment interaction style, such as single-agent Gym~\cite{brockman2016gym}, can effectively separate the environment from the algorithm, allowing the policy learner to remain oblivious to the details of the environment. To better favor the environment separation as well as facilitate future extension, we design a standardized multi-agent environment interaction wrapper. 
This wrapper establishes a consistent data style for multi-agent environments, so as to mitigate the operation inconvenience on different environments. Specifically, we encapsulate non-universal data, such as global state and action masks, into the \textit{obs}. The data within \textit{obs} is organized from the perspective of individual agents. The \textit{rewards} are structured as a set that includes rewards for two teams of agents. These rewards can be duplicated or allocated based on team rewards to each agent to ensure equity. In addition, the \textit{commands} interface allows the policy decisions to be executed within the environment. By employing the standardized interaction wrapper, APyMARL empowers algorithms to effectively utilize the data provided by various environments, thereby promoting flexibility and extensibility in the development/evaluation of MARL algorithms.  

\textbf{Dual-team adversarial training/testing controller}. APyMARL supports dual-team multi-agent combat in one scenario, where two teams of agents can learn policies themself through online competing interactions. Through the environment interaction wrapper, agents can independently access their own obs and rewards. According to the observation states, each team of agents could make joint policy decisions. Through using the \textit{commands} interface, the policy decisions are transmitted to the controller for updating the environment states. APyMARL can simultaneously load two different pre-trained algorithm-controlled models for real-time combat demonstrations. This can facilitate the development of dual-team multi-agent adversary research and expand the diversity of evaluation methods. 

\textbf{Parameter configurator}. A flexible and powerful configurator could make the library user-friendly. In the APyMARL library, one user can easily configure task and scenario parameters, algorithm hyperparameters, and adversarial paradigms~(supporting both single-team and dual-team paradigms) through the configurator.

In addition to the above features, APyMARL adopts a modular programming methodology and encapsulates many classic MARL techniques as modules for reusing or developing new algorithms. Moreover, it offers implementations of cutting-edge deep MARL algorithms that serve as benchmarks. Built upon the PyTorch framework, APyMARL facilitates rapid execution and training of deep neural networks while leveraging its extensive ecosystem.

\section{Experiments}


\subsection{Experiment Settings}

We perform several independent experiments (five runs) for each case and report their median/average performance by following the SMAC protocol. The primary evaluation metric is the median win percentage of evaluation episodes with different environment steps observed. Such metric can be estimated by periodically running a fixed number of testing episodes~(32 as default) without any exploratory behaviors. 

To reduce the effect of model initialization, we employ two methods for setting model parameters, random way as used in~\cite{he2015delving} and well-trained way~(trained with built-in AI paradigm). As the adversary between those state-of-the-art algorithms is often well-matched in strength, the dual-algorithm paired adversary mode needs a long-term process. Accordingly, we increase the number of training steps (10 million steps) to better discover the real ability of MARL algorithms. In the mixed adversary mode, opponent units exhibit a wider range of behaviors. We aim for algorithms to learn winning strategies through higher sampling efficiency rather than increasing the number of steps. Therefore, it is recommended to maintain the same number of training steps (2 million steps, same as SMAC) when evaluating subsequent algorithms. This ensures consistency in the evaluation process and allows for a fair comparison between different algorithms.

Based on our developed APyMARL, we conduct algorithm-vs-algorithm adversary tests, which refer to the eight representative algorithms: QMIX~\cite{rashid2020qmix}, VDN~\cite{sunehag2017vdn}, FOP~\cite{zhang2021fop}, DOP~\cite{wang2020dop}, QPLEX~\cite{wang2020qplex}, QTRAN~\cite{son2019qtran}, COMA~\cite{foerster2018coma} and IQL~\cite{tan1993iql}. 
In the dual-algorithm paired adversary mode, we aim to evaluate the algorithm's dynamic adversarial capabilities. This requires the combat scenarios to be identical layouts for both teams. Therefore, we conducted experiments mainly in symmetric scenarios. On the other hand, in the mixed adversary mode, we conducted experiments in all scenarios(both symmetric and asymmetric) for more comprehensive evaluation.

\subsection{Dual-algorithm Paired Adversary Experiments}
\label{smab_experimental_result}

\begin{figure}[!htb]
	\centering
	\subfloat{
		\includegraphics[scale=0.4]{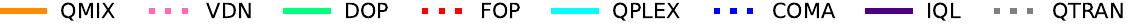}
	}
	
	\subfloat{
		\includegraphics[width=0.49\linewidth]{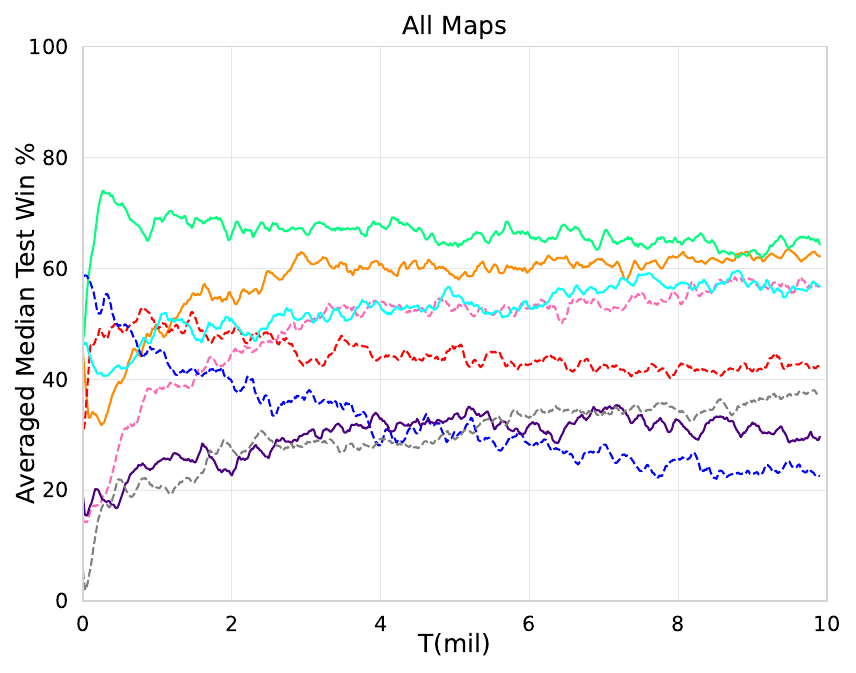}
	}
	\subfloat{
		\includegraphics[width=0.49\linewidth]{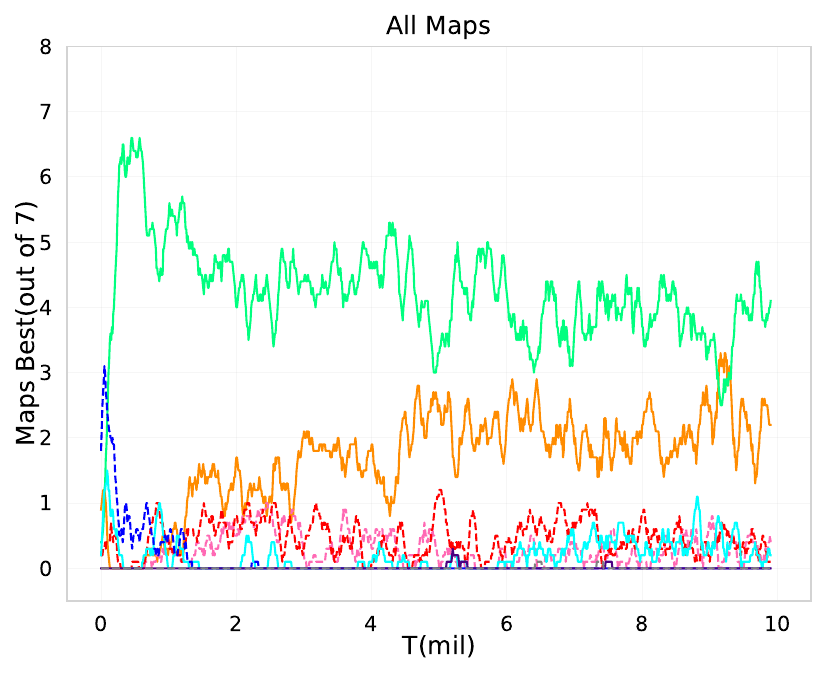}
	}
	\caption{The overall result of eight algorithms pairwise combat in seven symmetric scenarios under dual-algorithm adversary mode. Left: The median test win rates, averaged across all 7 scenarios. Right: The number of scenarios in which the algorithm outperform other algorithms~(median test win rate is highest by at least 1/32 and smoothed).}
	\label{sc2ba_paired_all}
\end{figure}

\begin{figure*}[!htb]
	\centering
	\subfloat{
		\includegraphics[scale=0.6]{picture/legend.pdf}
	}
	
	\subfloat{
		\includegraphics[width=0.245\linewidth]{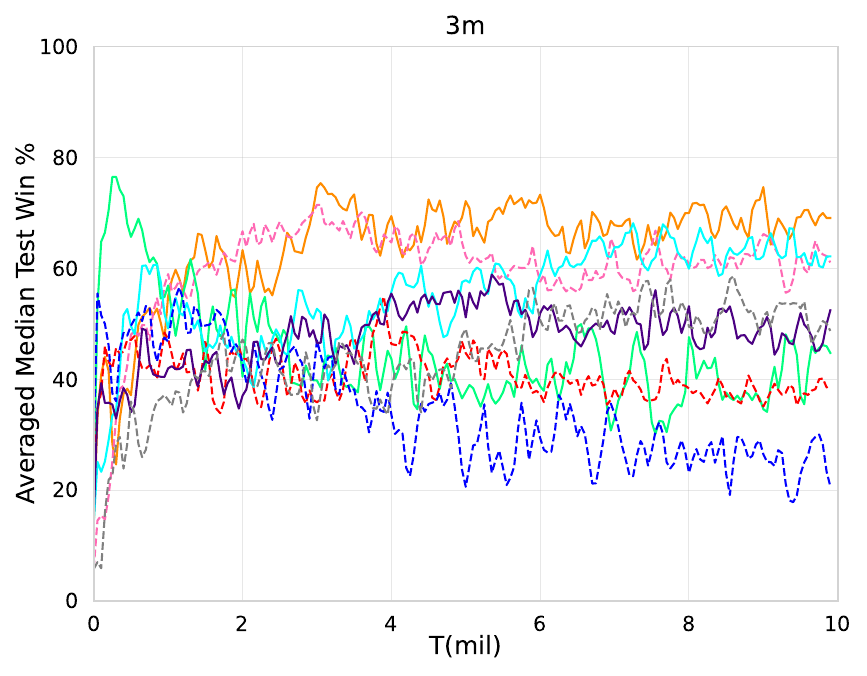}
	}
	\subfloat{
		\includegraphics[width=0.245\linewidth]{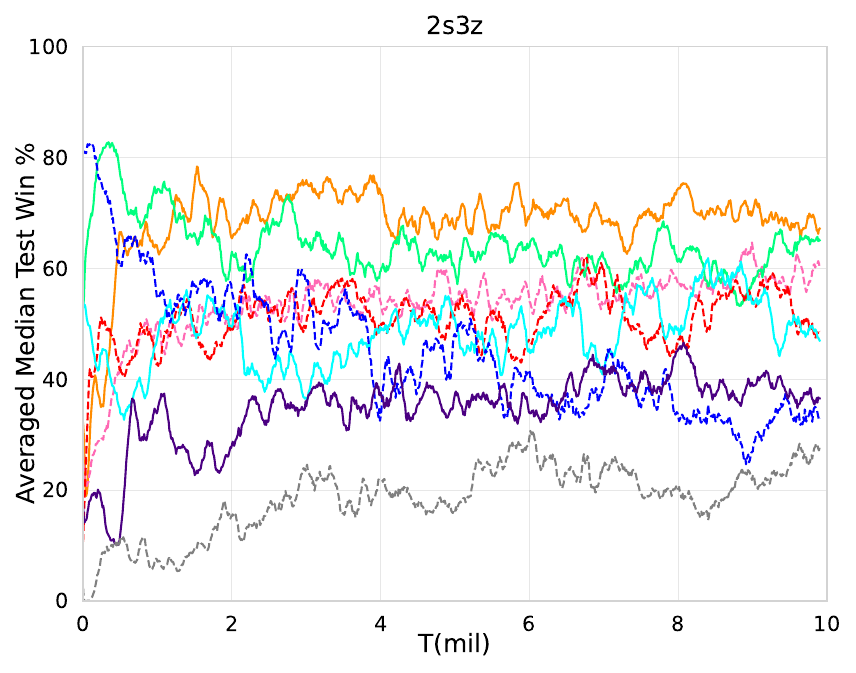}
	}
	\subfloat{
		\includegraphics[width=0.245\linewidth]{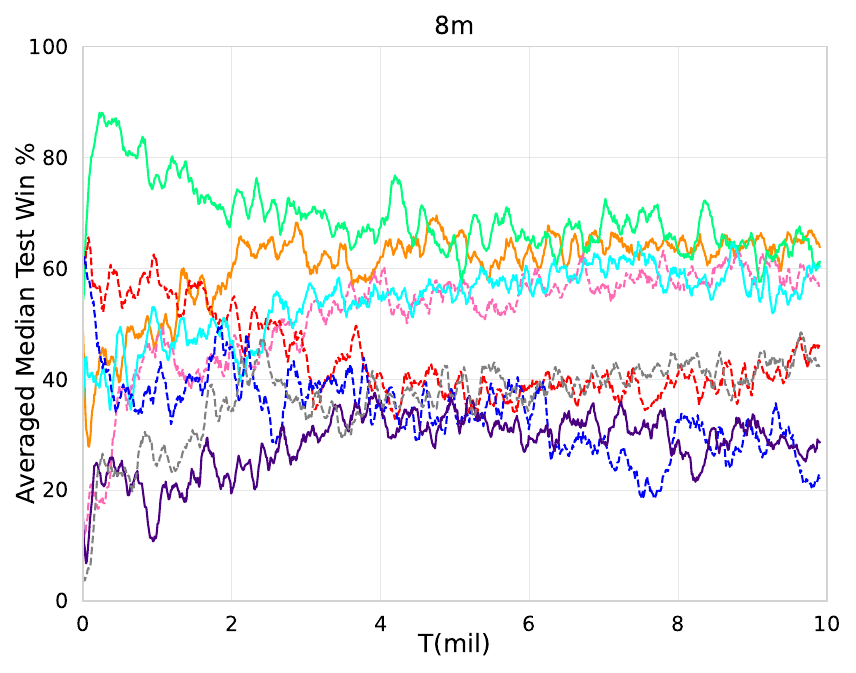}
	}
	\subfloat{
		\includegraphics[width=0.245\linewidth]{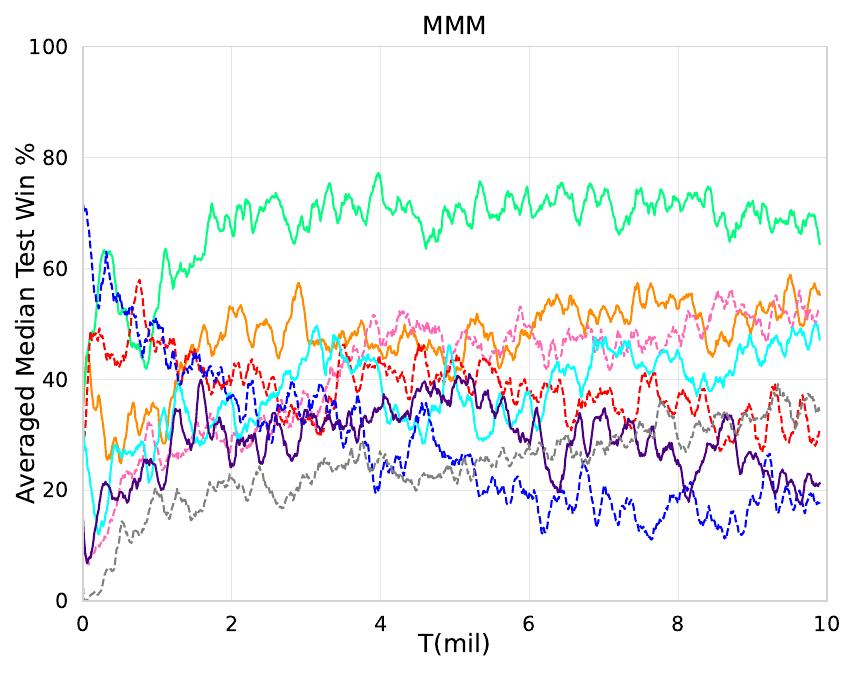}
	}
	
	\subfloat{
		\includegraphics[width=0.245\linewidth]{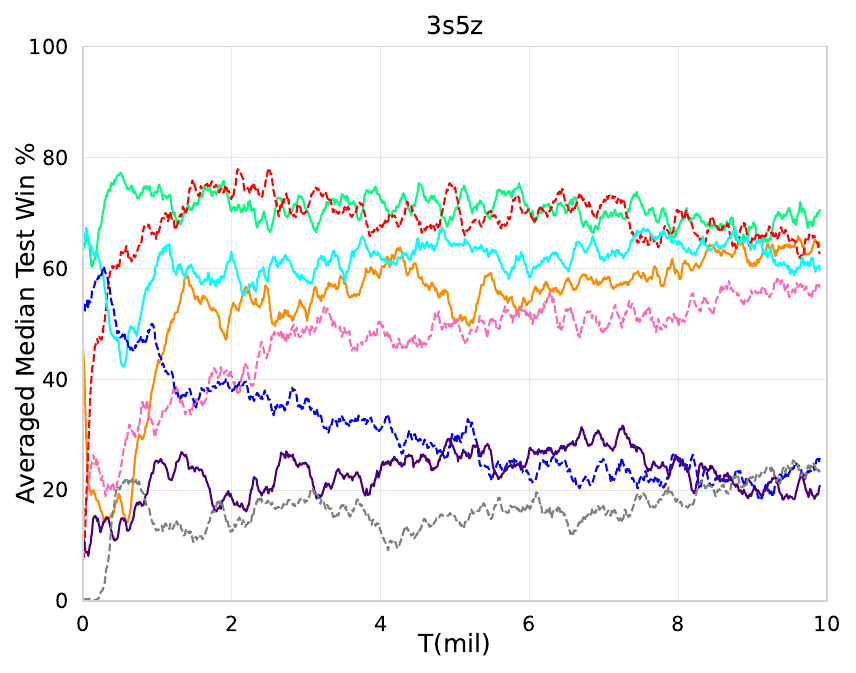}
	}
	\subfloat{
		\includegraphics[width=0.245\linewidth]{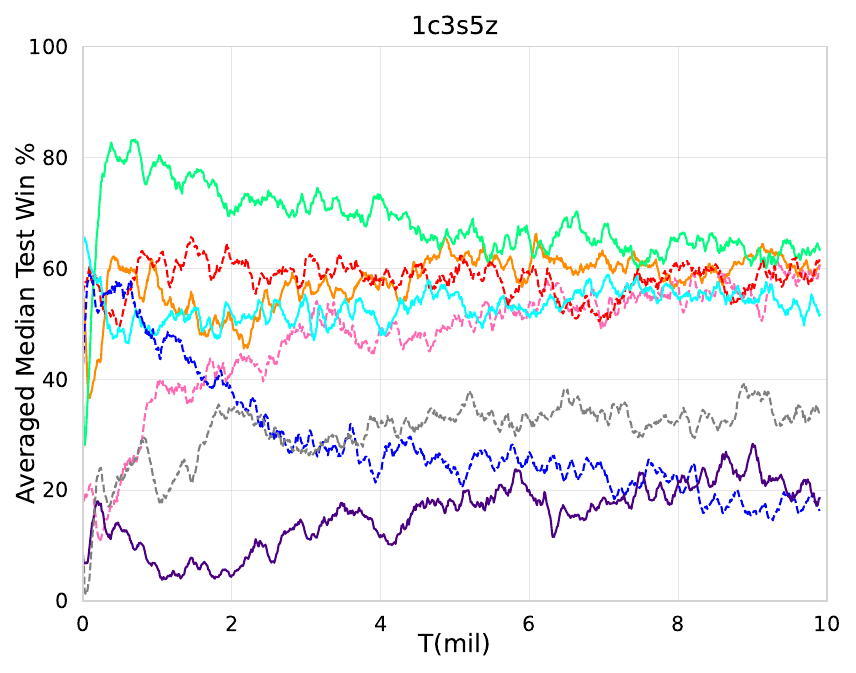}
	}
	\subfloat{
		\includegraphics[width=0.245\linewidth]{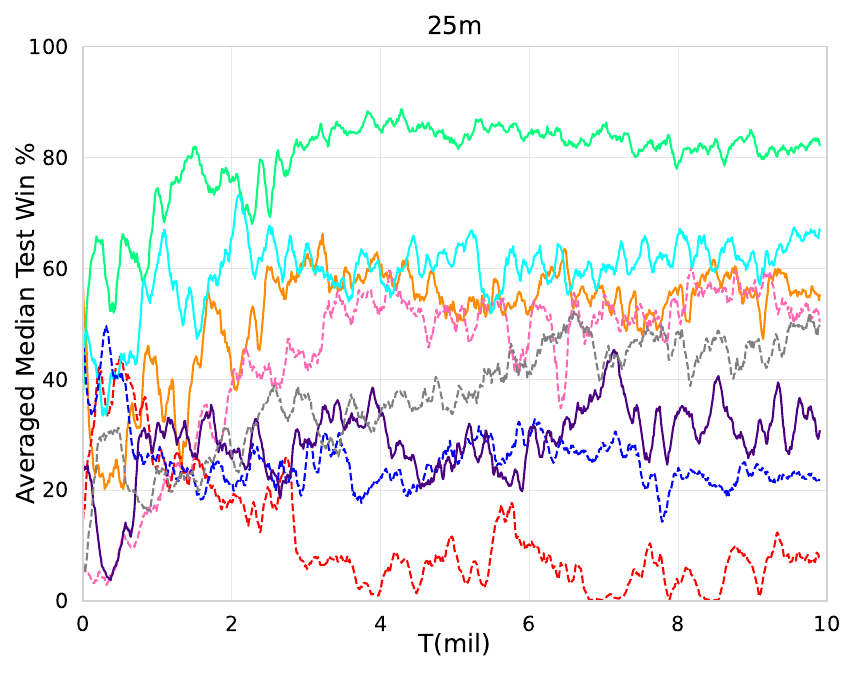}
	}
	\caption{Win rates for eight algorithms pairwise combat in seven symmetric scenarios(3m, 8m, 2s3z, 3s5z, MMM, 1c3s5z, and 25m) under dual-algorithm adversary mode. In one scenario, one algorithm(such as QMIX) will pairwise combat eight algorithms(include itself), then average eight win rates as the adversarial performance. Easy scenarios include 3m, 8m, and 2s3z, and hard scenarios consist of 3s5z, MMM, 1c3s5z, and 25m.}
	\label{sc2ba_paired_easy}
\end{figure*}

\begin{figure}[htb]
	\centering
	\subfloat{
		\includegraphics[scale=0.35]{picture/legend.pdf}
	}
	
	\subfloat{
		\includegraphics[width=0.49\linewidth]{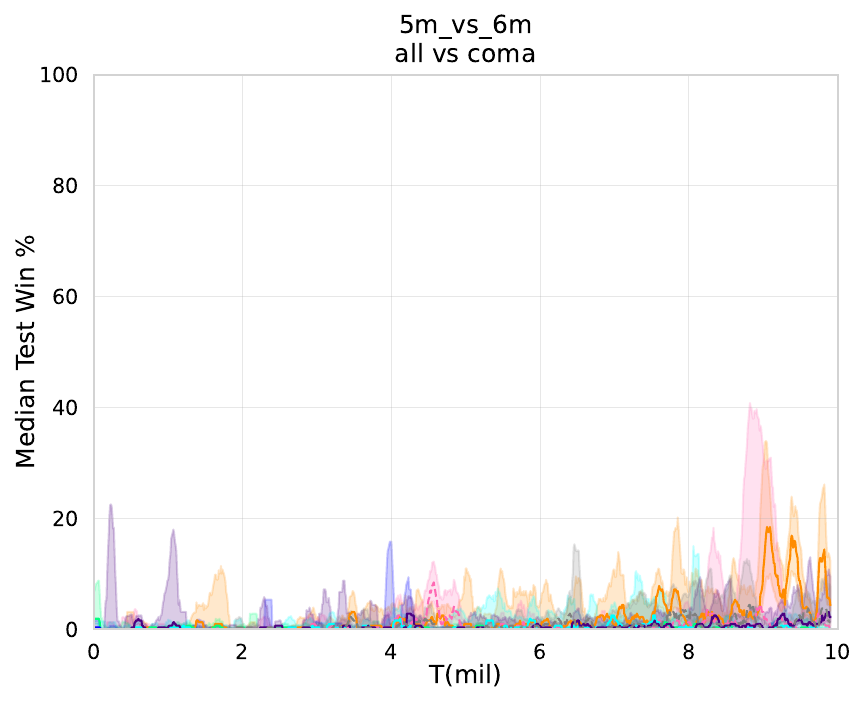}
		\includegraphics[width=0.49\linewidth]{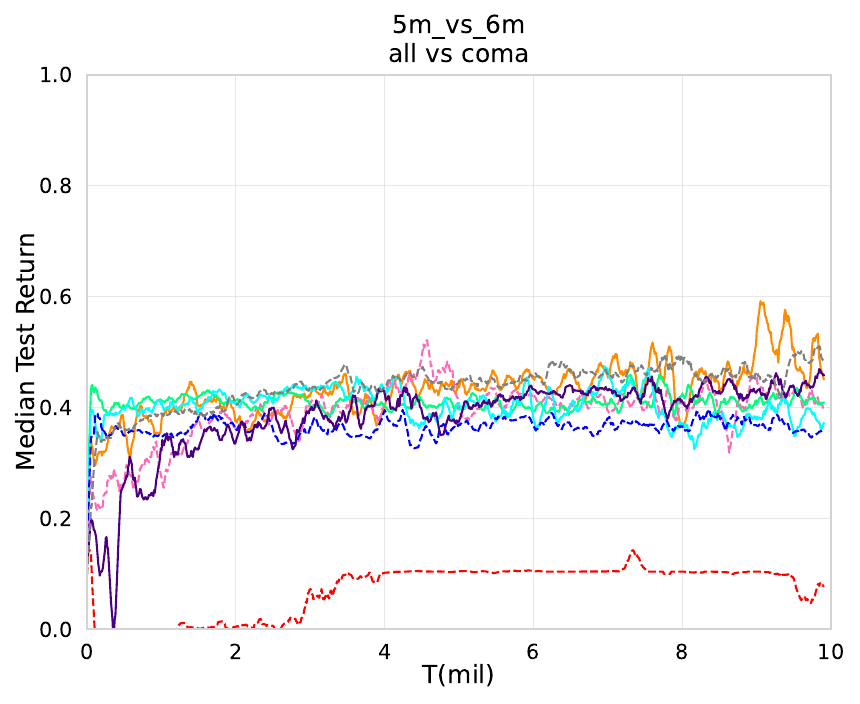}
	}
	\caption{The troop difference influence in dual-algorithm adversary mode. The results of all algorithms compete against the COMA in 5m\_vs\_6m are plotted here, including median win rate as well as median returns. The median returns are normalized to a range of 0-1.}
	\label{sc2ba_paired_5m6m}
\end{figure}

We compare the performance results of eight algorithms in the dual-algorithm paired adversary mode. To effectively evaluate the MARL algorithms' dynamic adversarial abilities, each algorithm is paired up against the others. The results are plotted in Fig.~\ref{sc2ba_paired_all}-Fig.~\ref{sc2ba_paired_5m6m}. \textit{The pairwise adversary of any two algorithms are deferred to the supplementary material.} Below we illustrate the detailed analysis.

\textbf{Overall performance}. The overall performance is measured based on two aspects: the average median win rate of all adversary cases, and the number of advantageous scenarios, as shown in Fig.~\ref{sc2ba_paired_all}. For the former, it reports the average value of the median win rates of the algorithm against all other algorithms~(including itself) in all scenarios. For the latter, it represents the number of scenarios in which the algorithm dominates. We can obtain some observations:

\begin{itemize}
	\item \textit{Dual-algorithm adversarial fluctuations}. In Fig.~\ref{sc2ba_paired_all}~(Left), the performances of all paired adversaries exhibit fluctuations. But in the popular SMAC environment controlled by AI bots, many algorithms can easily achieve nearly 100\% win rates~\cite{wang2020rode, wang2020roma, yu2022mappo, hu2024tvdo}. The reason should be that each algorithm endeavors to adjust for the opponent algorithm during the evolving game. In other words, one-team agents controlled by one algorithm need to continuously learn and adapt to dynamically changing opponents controlled by another algorithm. It also indicates that no fixed strategy can guarantee a consistent victory in the paired adversary mode.
	\item \textit{Performance difference in policy-/value-based methods}. The performance of policy-based methods seems less stable. Their performance tends to decline with the increase of steps, although they may gain certain advantages at the initial phase. Moreover, there are obvious performance discrepancies among these policy-based methods (DOP, FOP, and COMA), where DOP achieves the best performance. This can be attributed to its unique combination off-policy tree backup updates with the on-policy TD($\lambda$) technique, effectively addressing the challenge of lower training efficiency.
	In contrast, the value-based methods exhibit more stability. All algorithms basically show a consistent upward trend, indicating a gradually increasing ability adapting to dynamically changing opponents. For example, QMIX, VDN, and QPLEX make rapid improvements and eventually achieve performance close to DOP, although DOP initially gains an advantage. Additionally, these value-based methods perform at the same level, except for QTRAN. This can be attributed to the additional constraint, i.e., affine transformations, which makes it challenging for QTRAN to handle complex tasks~\cite{rashid2020qmix}.
	\item \textit{No algorithms dominate over all scenarios}. As shown in Fig.~\ref{sc2ba_paired_all}~(Right), both DOP and QMIX exhibit good performance, dominating in four and two scenarios respectively for the majority of phase. The other algorithms only occasionally achieve optimal performance in one scenario. No one (include best DOP) dominates all scenarios, which indicates the significant impact of scenario variations on algorithm performance. More discussion about the scenario influence is given below.
\end{itemize}

\textbf{Performance of different (symmetric) scenarios}. In these scenarios, each team consists of several agents ranging from 3 to 25. According to the types of units, these scenarios can be further classified into homogeneous and heterogeneous types. Homogeneous scenarios involve teams with agents of the same type, whilst heterogeneous scenarios feature teams with diverse types of units. The presence of multiple types of units introduces a rich variety of tactics in combat. Thus, algorithms need to effectively utilize different types of units, leveraging their strengths and avoiding weaknesses, to hold the upper hand. According to the number of agents and the diversity of agent types, we group the scenarios into \textit{easy} (3m, 8m and 2s3z) and \textit{hard} (MMM, 3s5z, 1c3s5z and 25m) categories. As shown in Fig.~\ref{sc2ba_paired_easy}, we can obtain some observations:

\begin{itemize}
	\item \textit{Comparable for best value-based and policy-based methods in easy scenarios}. According to the former three scenarios (3m, 8m, and 2s3z), the best ones of value-based and policy-based methods tend to be comparable in performance. For example, in the 8m and 2s3z scenarios, DOP (best for policy way) and QMIX (best for value way) reach an almost consistent performance at the terminate step. This can be attributed to the simplicity of the scenarios, where the cooperation policy is relatively straightforward, and the algorithm can learn effective policies more easily.
	\item \textit{More effective for policy-based way in hard scenarios}. As the quantity and variety of agents further increase, it poses challenges not only in terms of cooperative combat but also significantly expands the action space, leading to greater complexity in collaborative adversarial strategies.
	However, these policy-based methods are shown more effective and achieve promising results in hard scenarios, where DOP achieves best. Additionally, FOP performs well in the 3s5z and 1c3s5z scenarios, although it struggled in the 25m scenario, which might be mainly because of the optimal joint policy of entropy-regularized Dec-POMDP~\cite{oliehoek2016concise} rather than the original MDP, i.e., the converged policy may be biased~\cite{eysenbach2019maxent}. These policy-based methods, including DOP as well as FOP, are mostly superior to those value-based methods. It demonstrates that hard scenarios need be more driven by complicated policy learning compared to value-based learning.
	\item \textit{More performance discrepancy in hard scenarios}. As the complexity of scenarios increases, there are significant gaps in the performance of algorithms. In MMM and 25m scenarios, DOP outperforms other algorithms and establishes a substantial lead. In contrast, some algorithms perform poorly in all hard scenarios, as the learning of cooperative and competitive strategies becomes more complicated. Thus, the difficulty of scenarios will amplify the performance differences of algorithms. In other words, the increase of scenario complexity could be better used to evaluate the capability of various algorithms.
	\item \textit{More challenging in heterogeneous scenarios}. In multiple types of unit scenarios, algorithms need to consider collaboration among various types of agents that possess distinct attributes/actions. When comparing the 8m (homogeneous) and 3s5z (heterogeneous) scenarios, both featuring 8 agents in one team, there is a greater disparity in algorithm performance, which has two forks with either most success or most failure.  
	It shows that the cooperation tactics of different types of agents pose challenges for algorithms. For this, some works~\cite{wang2020rode, wang2020roma} employed some specific collaborative strategies for this heterogeneous case.
	\item \textit{The effect of the quantity of agents}. In the 3m scenario, all algorithms demonstrate strong adversarial capabilities, with most algorithms achieving an average win rate distributed in a narrow range. In contrast, in the 8m and 25m scenarios where only the number of agents is increased, there are significant differences in algorithm performance. The results indicate that the increased quantity of agents presents challenges for algorithms to learn effective strategies, leading to significant performance gaps among some algorithms.
	
\end{itemize} 

\textbf{Asymmetric scenarios}. We also conduct experiments on asymmetric scenarios. The asymmetry with slight troop difference might easily overwhelm an advantage algorithm. To verify this point, we select the 5m\_vs\_6m scenario and let all algorithms compete against the COMA performing the poorest in experiments. The win rates and episode returns are depicted in Fig.~\ref{sc2ba_paired_5m6m}, from which we can obtain some observations:

\textit{MARL Algorithms are extremely sensible to (even slight) troop changes}. Only less an agent (five agents vs six agents), as shown in Fig.~\ref{sc2ba_paired_5m6m}~(Left), all algorithms (including the best DOP) almost failed when competing with the algorithm COMA (the poorest in the above symmetric experiments). For the episode rewards in Fig.~\ref{sc2ba_paired_5m6m}~(Right), all algorithms do not exceed 0.6, and only QMIX occasionally makes slightly progress in the final stages. It demonstrates that the asymmetric setting increases the dual-algorithm adversary difficulty, and the current MRAL algorithms at least lack scalability in the troop layout. Consequently, it indicates some new strategies, such as rescheduling reward function, are required to encourage MARL algorithms to overcome the bottleneck incurred by troop layout.

\textit{In summary}, for the algorithm-paired adversary mode, we have some conclusions:
\begin{itemize}
	\item \textit{The dual-algorithm adversary provides another measurement mode of evaluating MARL algorithms}. Different from the Built-in AI Bots mode that evaluates algorithm performance through battles against fixed strategy bots, this mode offers an online adversarial setting to compete with another MARL algorithm. In this new mode, the policy of opponents is dynamic and evolvable under the control of MARL algorithm. Consequently, no fixed strategy can guarantee consistent victory, and agents need more powerful adaptation to evolving opponents.
	\item \textit{The complicated scenarios are more challenging to learn cooperation strategies}. As the number of agents and the diversity of agent types increase, learning cooperative strategies becomes more complex. This would lead to amplified performance differences among algorithms, while imposing higher demands on them. A specific example is the significant gap in algorithm performance observed in hard scenarios. It indicates that the increase of scenario difficulty can benefit for evaluating the adversarial capability of algorithms, allowing outstanding algorithms to well demonstrate their strengths.
	\item \textit{The slight difference of troops easily results in the failing of an advantage algorithm}. When dual-algorithm combat in one scenario, even a little disadvantage can lead to a complete failure. In unequal troops combat scenarios, even excellent algorithms may struggle to overcome an enemy with both troop advantages and evolutionary capabilities. For example, in the 5m\_vs\_6m scenario, the best algorithm DOP can not yet defeat the poorest algorithm COMA when COMA only adds an agent unit. It indicates that to overcome this bottleneck, new algorithms/techniques need to be developed, including rescheduling reward mechanisms, enriching evaluation metrics, etc. 
\end{itemize}

\subsection{Multi-algorithm Mixed Adversary Experiments}
\label{sec:mixed_adversary_mode}
\begin{figure}[htbp]
	\centering
	\subfloat{
		\includegraphics[scale=0.4]{picture/legend.pdf}
	}
	
	\subfloat{
		\includegraphics[width=0.49\linewidth]{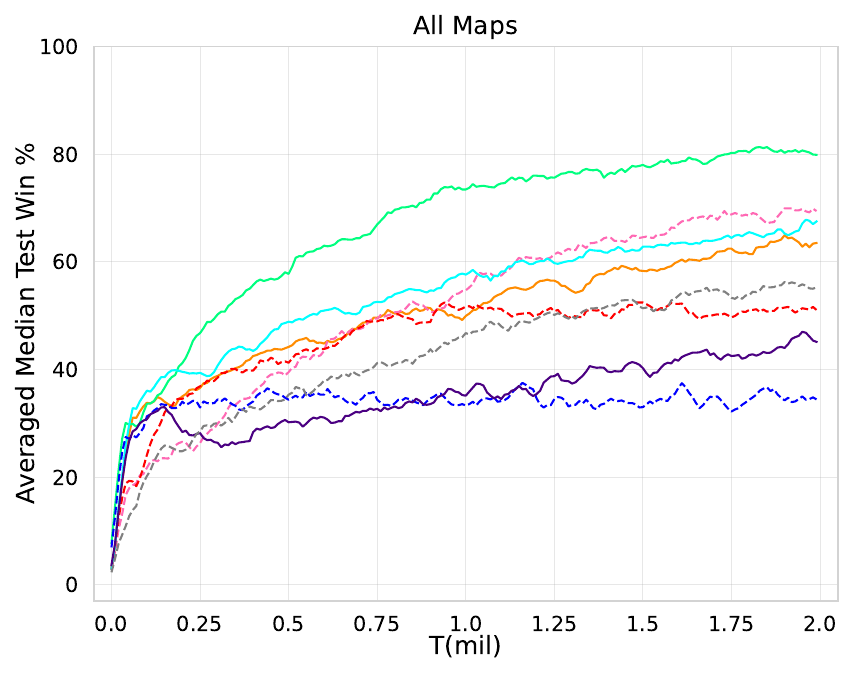}
	}
	\subfloat{
		\includegraphics[width=0.49\linewidth]{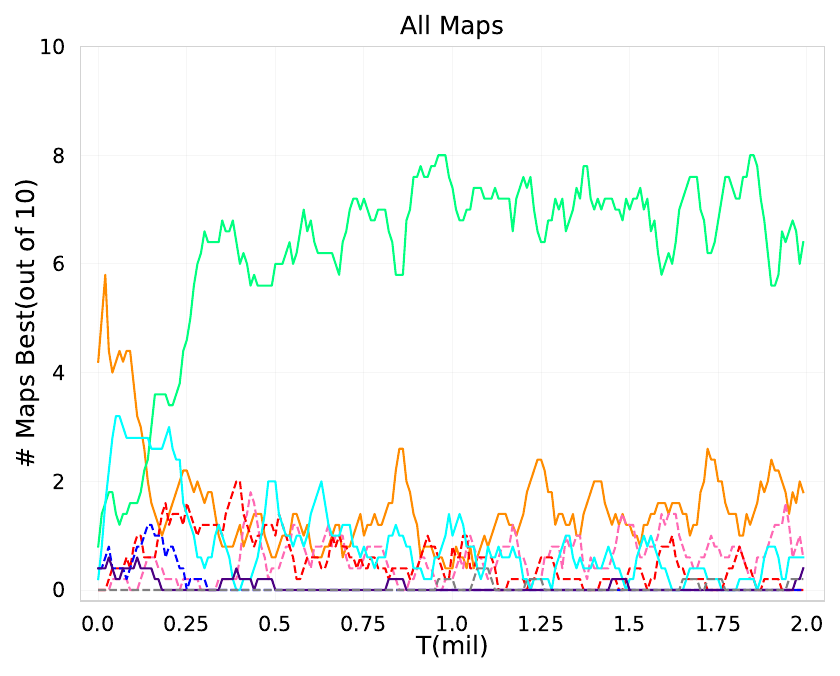}
	}
	\caption{The overall result of eight algorithms in multi-algorithm mixed adversary mode. Left: The median test win rates, averaged across all 10 scenarios. Right: The number of scenarios in which the algorithm outperforms other algorithms~(median test win rate is highest by at least 1/32 and smoothed).}
	\label{sc2ba_multi_all}
\end{figure}

\begin{figure*}[!htbp]
	\centering
	\subfloat{
		\includegraphics[scale=0.6]{picture/legend.pdf}
	}
	
	\subfloat{
		\includegraphics[width=0.245\linewidth]{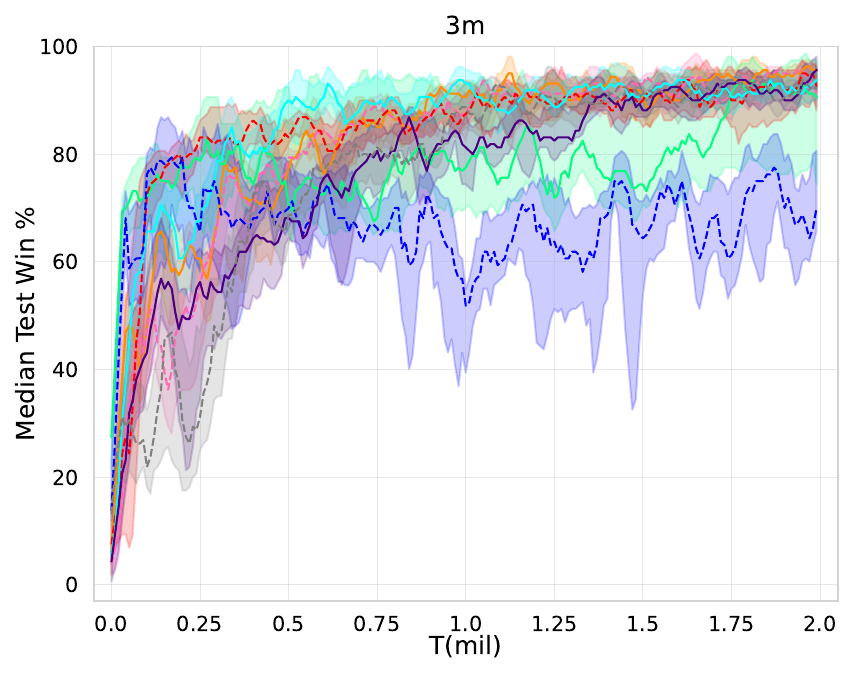}
	}
	\subfloat{
		\includegraphics[width=0.245\linewidth]{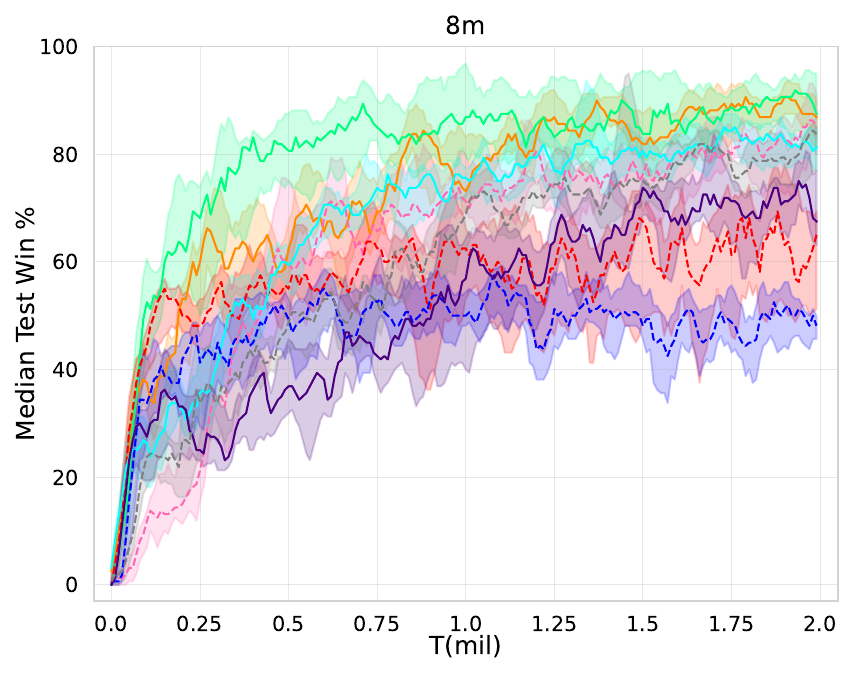}
	}
	\subfloat{
		\includegraphics[width=0.245\linewidth]{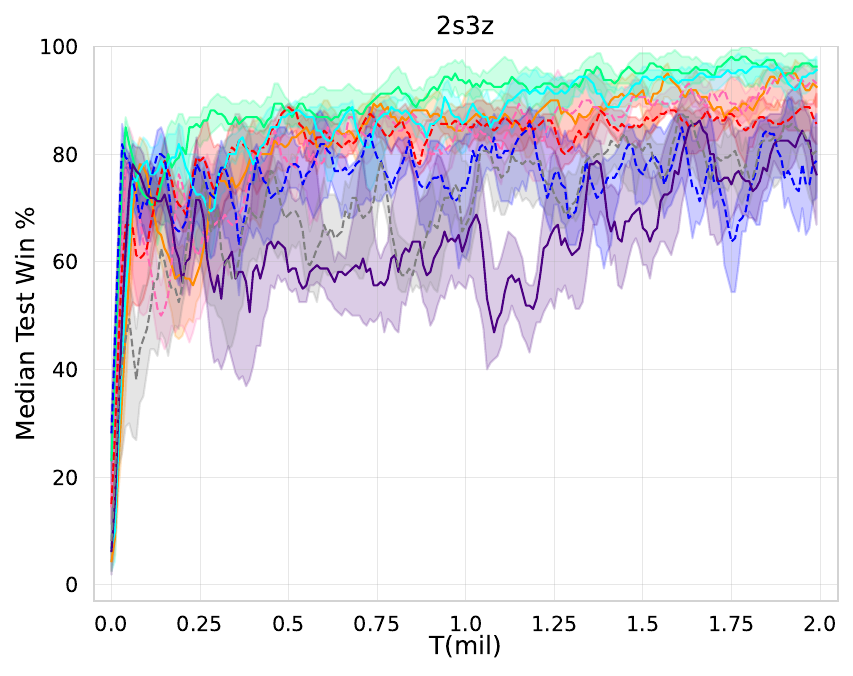}
	}
	\subfloat{
		\includegraphics[width=0.245\linewidth]{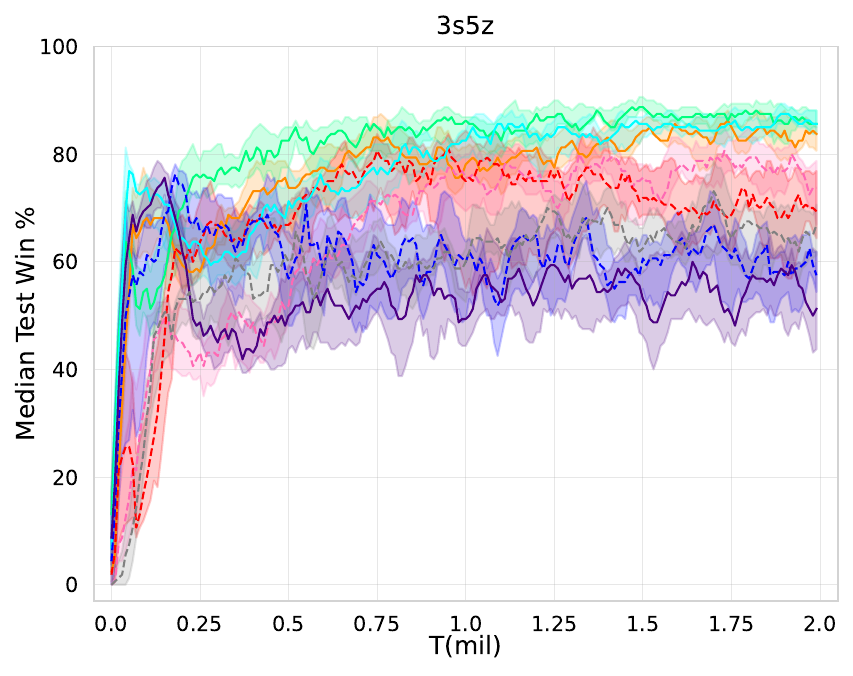}
	}
	
	\subfloat{
		\includegraphics[width=0.245\linewidth]{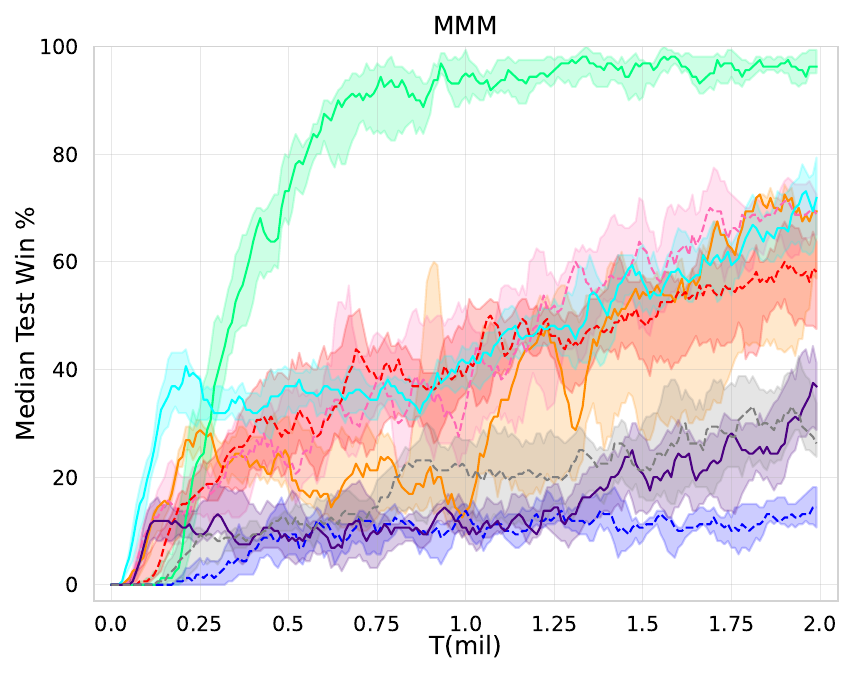}
	}
	\subfloat{
		\includegraphics[width=0.245\linewidth]{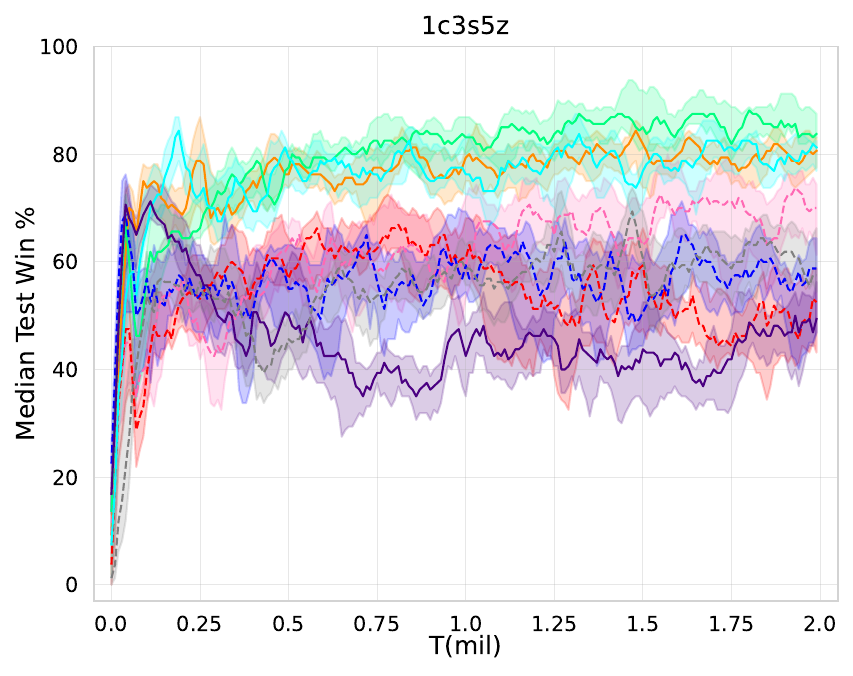}
	}
	\subfloat{
		\includegraphics[width=0.245\linewidth]{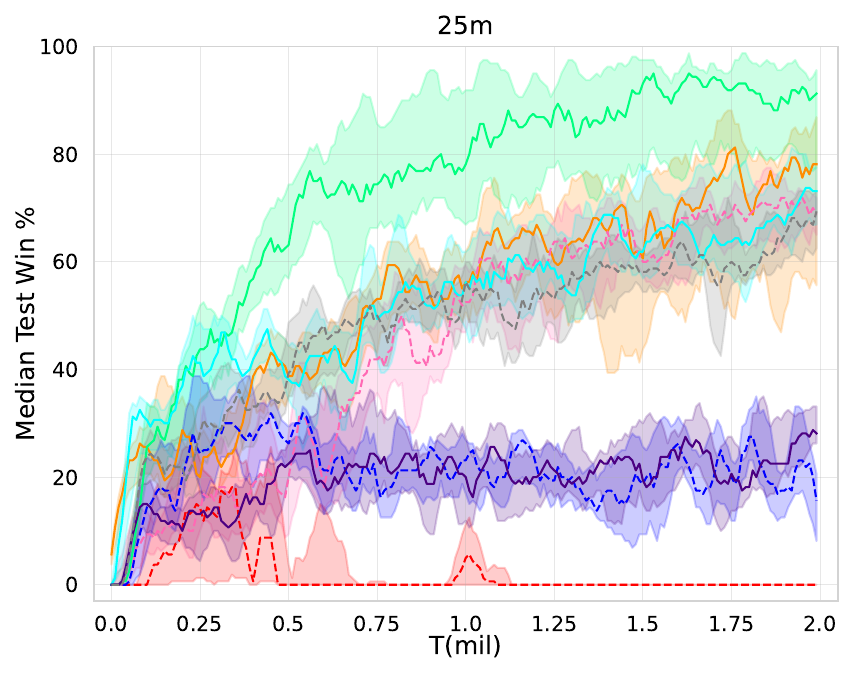}
	}
	\caption{Win rates for eight algorithms combat multi pre-trained models in seven symmetric scenarios(3m, 8m, 2s3z, 3s5z, MMM, 1c3s5z, and 25m) under multi-algorithm mixed adversary mode. }
	\label{sc2ba_multi_eazy}
\end{figure*}

\begin{figure*}[!htbp]
	\centering
	\subfloat{
		\includegraphics[scale=0.6]{picture/legend.pdf}
	}
	
	\subfloat{
		\includegraphics[width=0.245\linewidth]{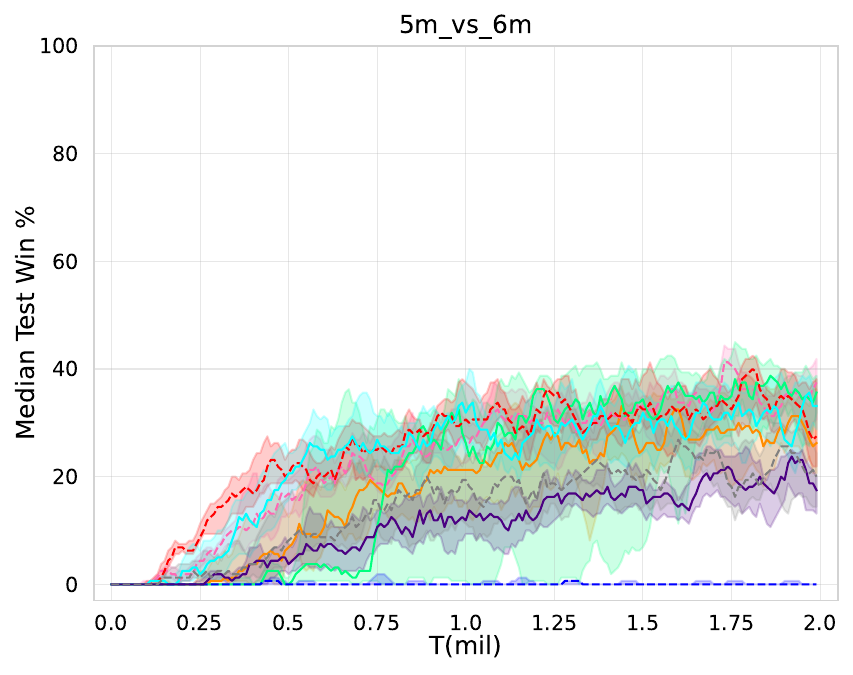}
	}
	\subfloat{
		\includegraphics[width=0.245\linewidth]{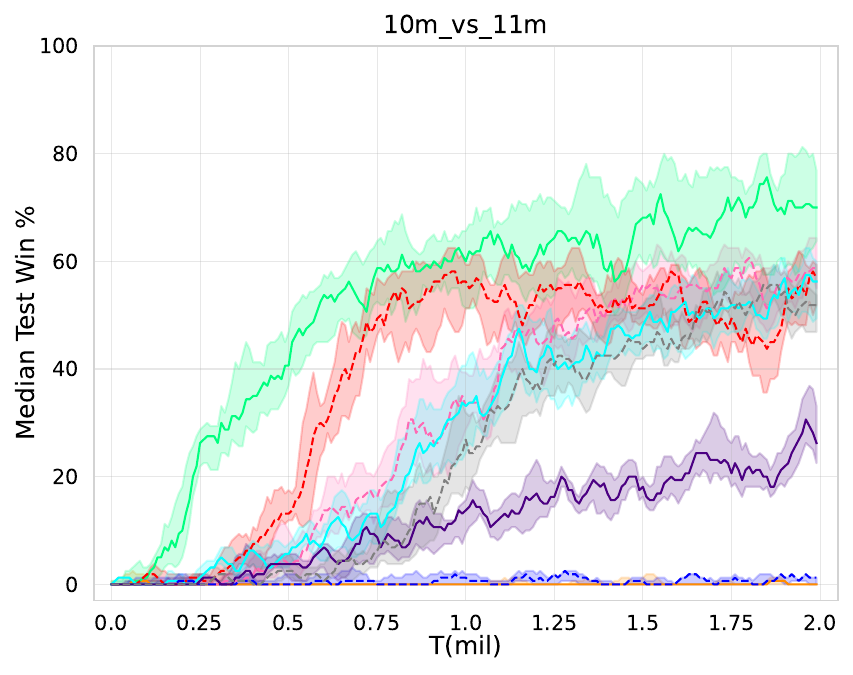}
	}
	\subfloat{
		\includegraphics[width=0.245\linewidth]{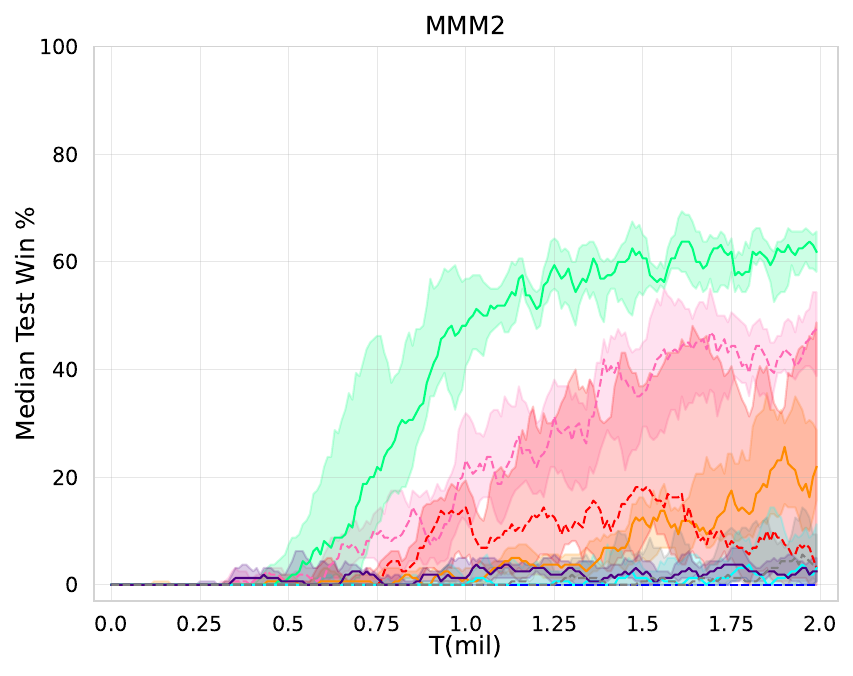}
	}
	\caption{Win rates for eight algorithms combat multi pre-trained models in three asymmetric scenarios(5m\_vs\_6m, 10m\_vs\_11m, and MMM2) under multi-algorithm mixed adversary mode.}
	\label{sc2ba_multi_supper}
\end{figure*}

We compare the performance results of eight algorithms in the mixed adversary mode. In these scenarios, one algorithm competes with the nine pre-trained and then fixed models (i.e., eight algorithm models and one built-in AI bot). Enemy units are controlled alternately through the nine models to increase the diversity of policies. Please note that these models have been pre-trained to master various combat techniques, such as focus fire, but they are not updated during the adversary phase due to the optimization complexity, which is also an open issue beyond this work.
The results are reported in Fig.~\ref{sc2ba_multi_all}-Fig.~\ref{sc2ba_multi_supper}. Below we illustrate more details analysis.

\textbf{Overall performance}. The average values of the median win rates of one algorithm against in all scenarios~(both symmetric and asymmetric) are shown in Fig.~\ref{sc2ba_multi_all}, we can have some observations:

\begin{itemize}
	\item \textit{Ascending trends in win rates with smaller fluctuations (vs dual-algorithm adversary)}. In Fig.~\ref{sc2ba_multi_all}~(Left), all algorithms generally exhibit consistent ascending trends in performance. When compared to dual-algorithm adversaries, the mixed adversaries perform more stable (i.e., less fluctuations). The reason might be that, the opponent is diverse (alternate eight algorithms), but not be dynamically updated due to the difficulty of optimization (also an open issue beyond this work). In other words, the mixed adversary mode offers a more stable adversary environment, allowing algorithms to continuously accumulate their experience to adapt to the opponents. For instance, as the number of training steps increases, DOP's performance continues to improve and eventually reaches a win rate of 80\%.
	\item \textit{Better consistency in performance for different value-based algorithms}. 
	The value-based algorithms almost have more consistency improvement compared to the policy-based methods. For instance, while QTRAN initially performs lower than FOP at the early stage, it consistently improves over time and eventually surpasses FOP. Furthermore, the value-based algorithms have better performance than those policy-based algorithms except for DOP. This indicates the stableness of value-based methods, which can be attributed to their higher computational efficiency, as the estimation and updating of value functions can be performed iteratively.
\end{itemize}

\begin{figure*}[!htbp]
	\centering	
	
	\includegraphics[scale=0.8]{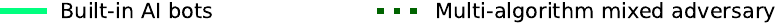}
	\subfloat[The DOP]{
		\includegraphics[width=0.245\linewidth]{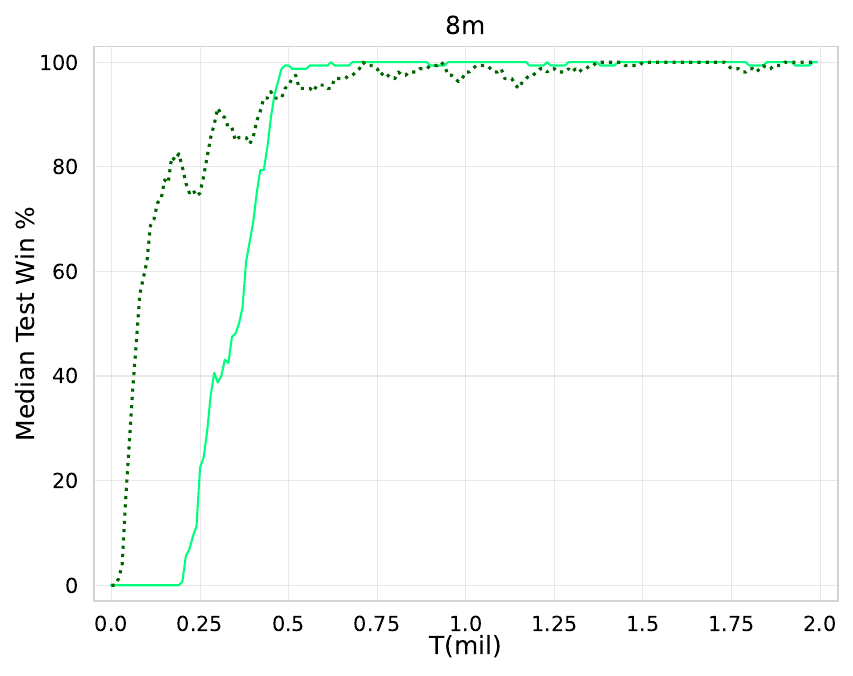}
		\includegraphics[width=0.245\linewidth]{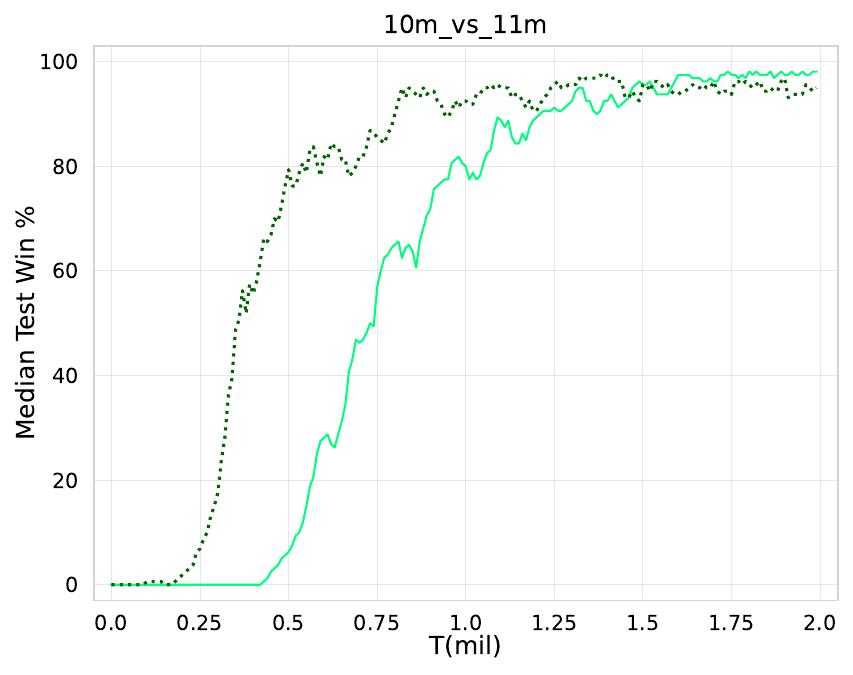}
		\includegraphics[width=0.245\linewidth]{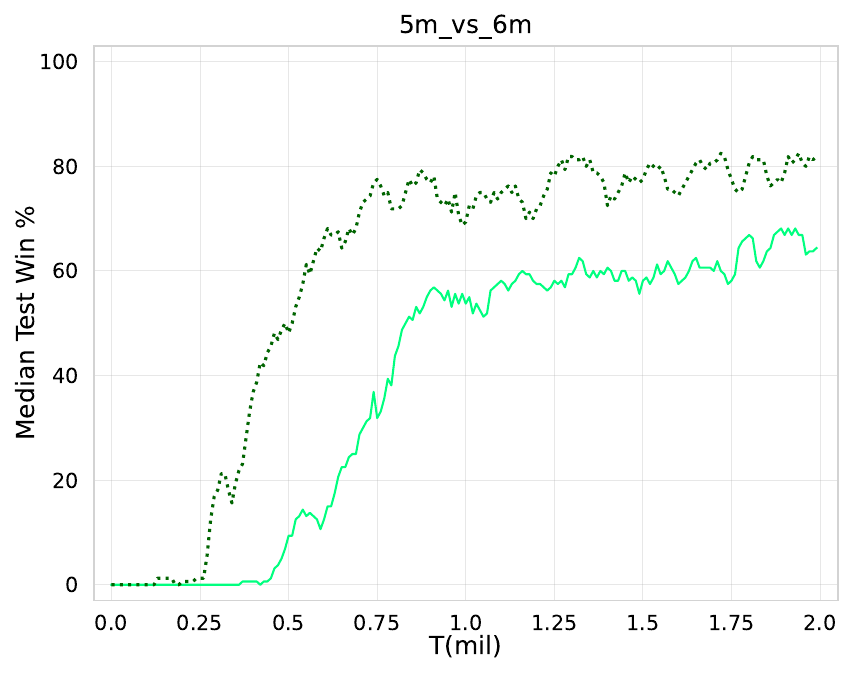}
		\includegraphics[width=0.245\linewidth]{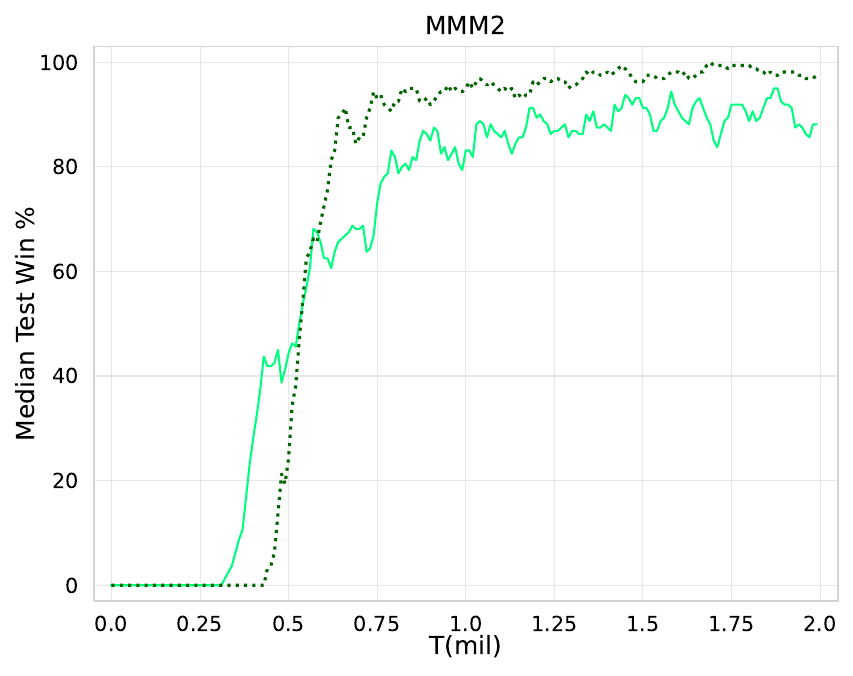}
	}
	\hspace{2cm}
	
	\includegraphics[scale=0.8]{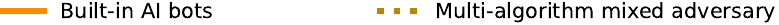}
	\subfloat[The QMIX]{
		\includegraphics[width=0.245\linewidth]{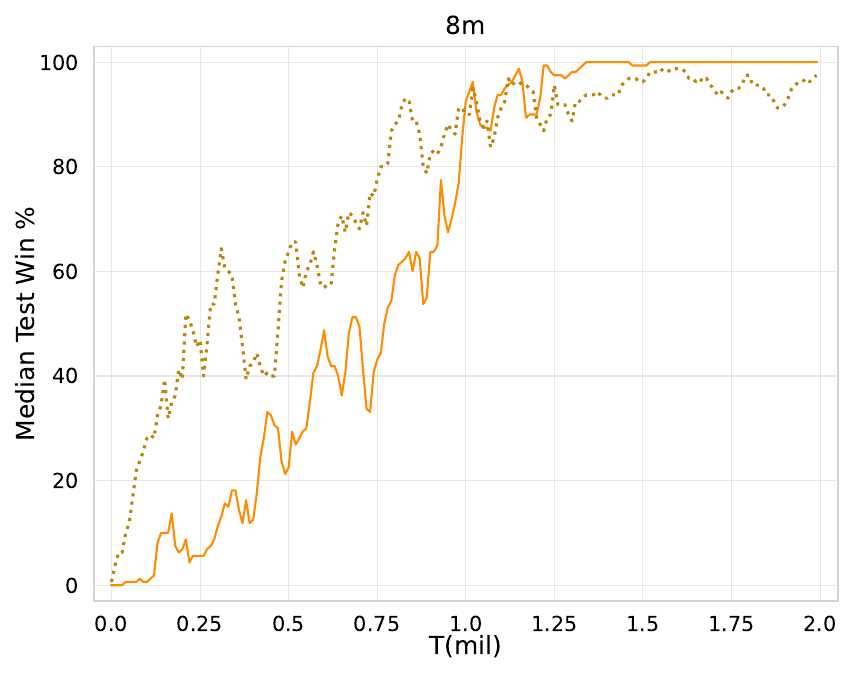}
		\includegraphics[width=0.245\linewidth]{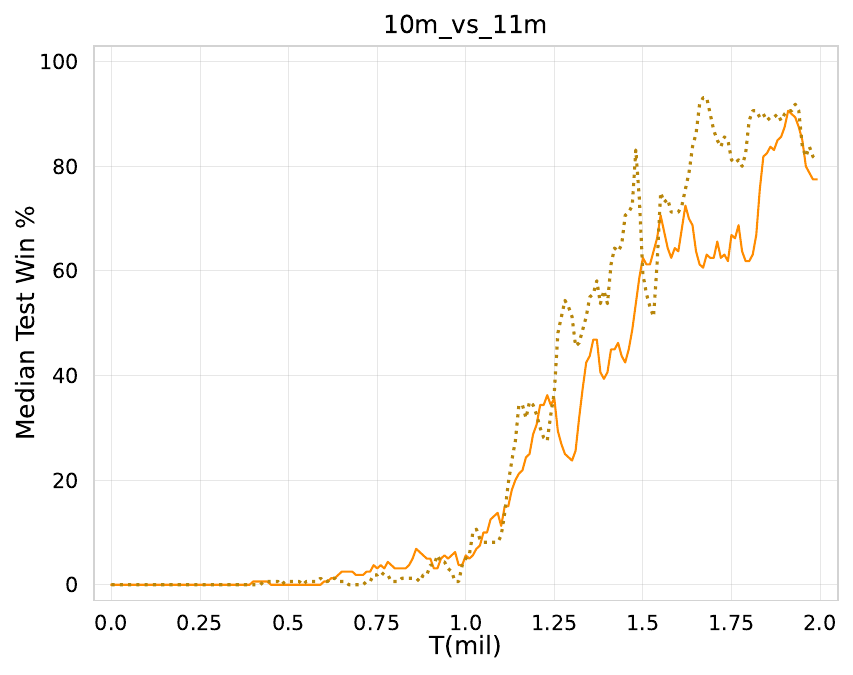}
		\includegraphics[width=0.245\linewidth]{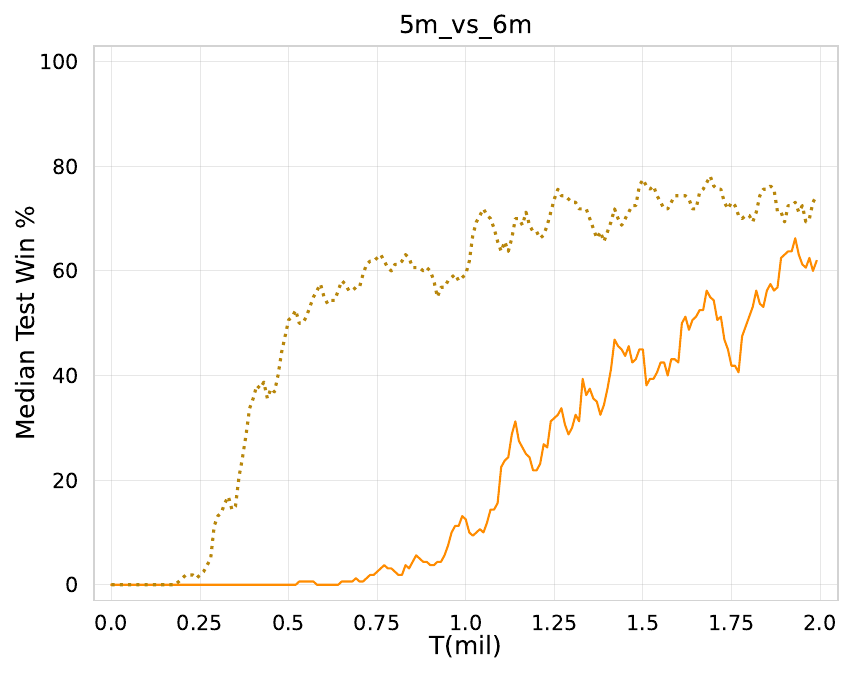}
		\includegraphics[width=0.245\linewidth]{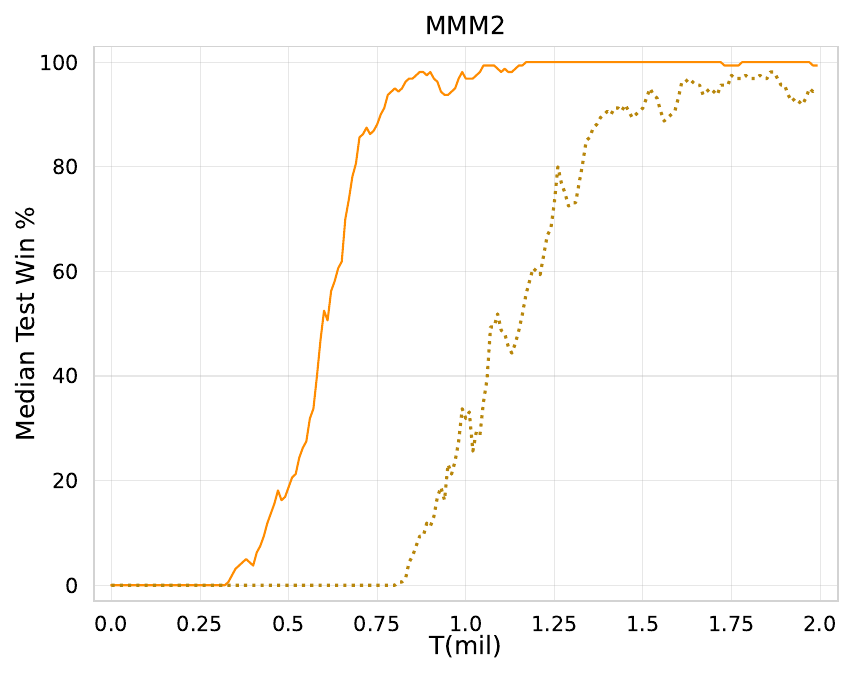}
	}
	
	\caption{The performance adversarial to built-in AI bots of algorithms under two training mode(built-in AI bots mode and multi-algorithm mixed adversary mode) in four scenarios(8m, 10m\_vs\_11m, 5m\_vs\_5m and MMM2). DOP and QMIX are trained in two mode respectively, and the win rate plotted here are obtained from battles against built-in AI bots. Please see the detail in Section~\ref{sec:opponents diversity}}
	\label{improve_opponent_diverse}
\end{figure*}

\textbf{Symmetric scenarios}. These scenarios consist of seven unique cases in which two camps share an identical layout. The combat results are shown in Fig.~\ref{sc2ba_multi_eazy}, from which we can observe that one-vs-multiple adversary is a challenging task even in some simple scenarios. This indicates that increasing the diversity of opponents can prevent algorithms from exploiting some simple policies for victory. But we can find that such mixed opponents could boost the generalization of one algorithm as discussed in Section~\ref{sec:opponents diversity}. 
Furthermore, in some scenarios that require more complex policies, such as MMM and 25m, some algorithms such as COMA and IQL, do not reach the win rate of 40\%. It demonstrates the challenges posed by the mixed adversary mode, where diverse opponents require algorithms to have higher sampling efficiency.

\textbf{Asymmetric scenarios}. We conduct experiments from two folds: i) different troop strengths but with homogeneous agent units, 5m\_vs\_6m (plus 20\% troop) and 10m\_vs\_11m (plus 10\% troop); ii) heterogeneous agent units, the MMM2 scenario. In these scenarios, the red-team agents are placed in unfavorable conditions. 
Agents not only have to confront opponents with diverse policy actions but also need to overcome disadvantages in terms of troop strength. The combat results are shown in Fig.~\ref{sc2ba_multi_supper}, from which we can obtain some observations. 
Firstly, this setting poses a significant challenge for MARL algorithms, none of the algorithms can achieve the same level of performance as in symmetric scenarios. 
Secondly, the slight variation of troop strength has a significant impact on the algorithm adversary. Comparing the scenarios of 5m\_vs\_6m (plus 20\% troop) and 10m\_vs\_11m (plus 10\% troop), the larger disparity in troops makes the 5m\_vs\_6m scenario more difficult to overcome, with none of the algorithms achieving a win rate higher than 50\%. 
Thirdly, in the scenario MMM2 of heterogeneous agent units, we observe substantial differences in algorithm adversary from 5m\_vs\_6m. Despite facing the same troop strength disadvantage of 20\%, DOP and VDN outperform other algorithms in MMM2. 
That might be attributed to diverse types and larger quantities of units introducing greater pool capacity of available tactical actions. Consequently, this presents an opportunity to develop more complex policies to effectively overcome unfavorable settings.

\textit{In summary}, for multi-algorithms mixed adversary mode, we have the conclusions:

\begin{itemize}
	\item \textit{The mixed adversary mode offers a powerful way to boost MARL performance}. 
	In this mode, we incorporated multiple well-trained enemy control models to diversify the behavior patterns of opponents. The increasing opponents diversity would force agents to adapt to different opponents, which boosts the red-team agents to master a wider range of tactical policies. Please see the discussion in Section~\ref{sec:opponents diversity}.
	\item \textit{Asymmetric scenarios are still challenging for the current algorithms}. In the mixed algorithm adversary, the algorithm is still sensible to the slight equality of troops, as also observed in the above dual-algorithm adversary experiments mentioned earlier. In the real world, it is unrealistic to expect equal troop strength. Hence, some new algorithms/techniques need to be developed urgently to address the case. 
	\item \textit{Dynamic mixed adversary is more complicated as the future work}. 
	In the above mixed adversary experiments, we fix the multiple blue-team models after well pre-trained, and then alternately choose one model as an opponent for each episode. Although the experiment demonstrates its effectiveness, there is still space to improve the fixed mode by jointly optimizing the multiple blue-team models, as used in the dynamic update of the dual-algorithm adversary mode. However, the joint multi-model optimization is destined to be complicated, and will be studied as a future work.
\end{itemize}

\subsection{Boost Multi-agent Learning}

\begin{figure*}[!htbp]
	\centering	
	\includegraphics[scale=0.7]{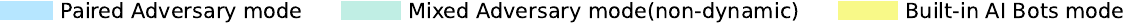}
	\subfloat{
		\includegraphics[width=0.33\linewidth]{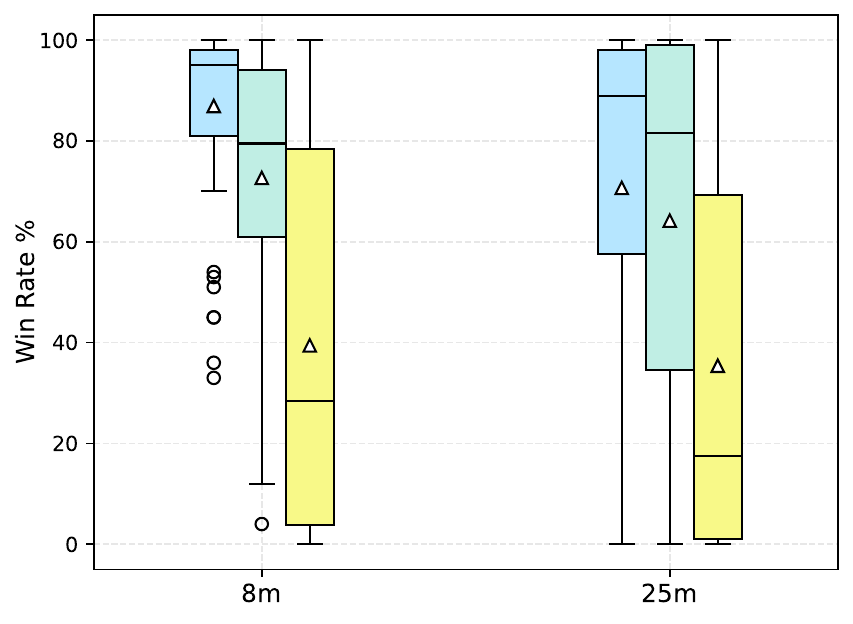}
	}
	\subfloat{
		\includegraphics[width=0.33\linewidth]{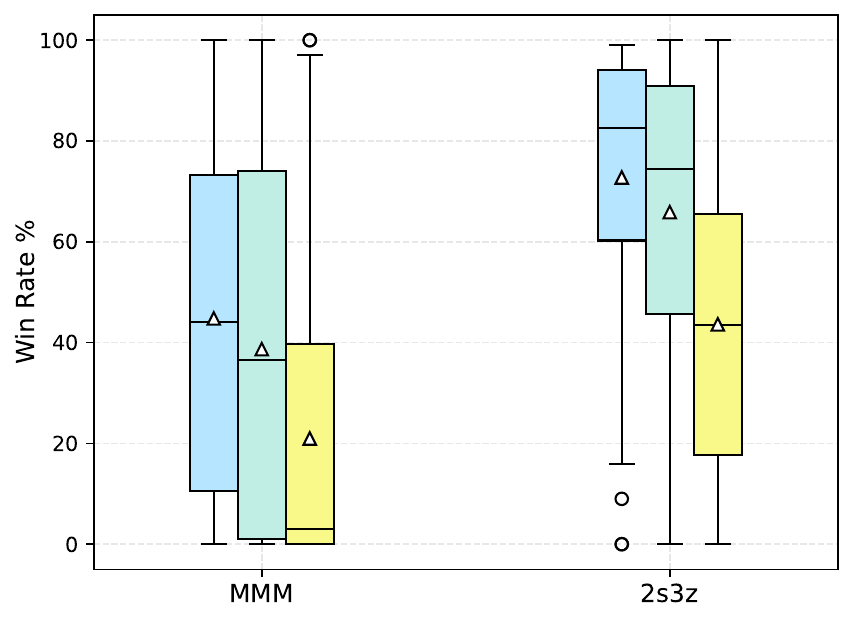}
	}
	\subfloat{
		\includegraphics[width=0.33\linewidth]{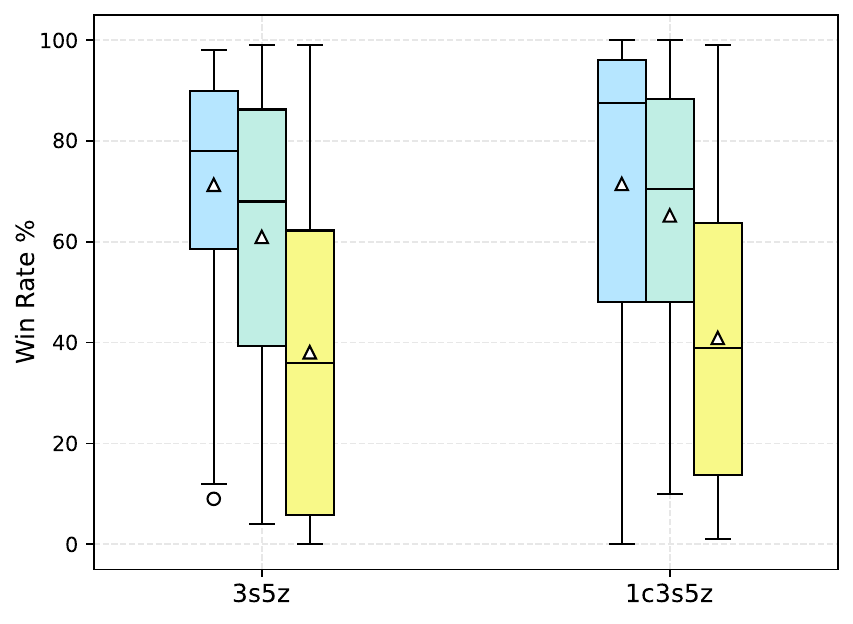}
	}
	\caption{The adversary ability of algorithms under three training modes (paired adversary mode, mixed adversary mode and built-in AI bots). The eight MARL algorithms are trained based on the three modes respectively, and reach 24 models for each scenario. Each model from the former two modes is competed to everyone from built-in AI bots, i.e., cross-adversarial tests in the one-to-one manner. The average results of all adversarial tests are plotted here. 
		In each box, the center/bottom/top lines represent the median, first quartile and third quartile, respectively; the horizontal lines along the top and bottom edges represent the maximum and minimum values; and the white triangle denotes the average value. Please see the details in Section~\ref{sec:adversary_competing}. }
	\label{multi_paired_generalize}
\end{figure*}

\begin{figure*}[!htbp]
	\centering	
	\subfloat[Built-in AI bits]{
		\includegraphics[width=0.33\linewidth]{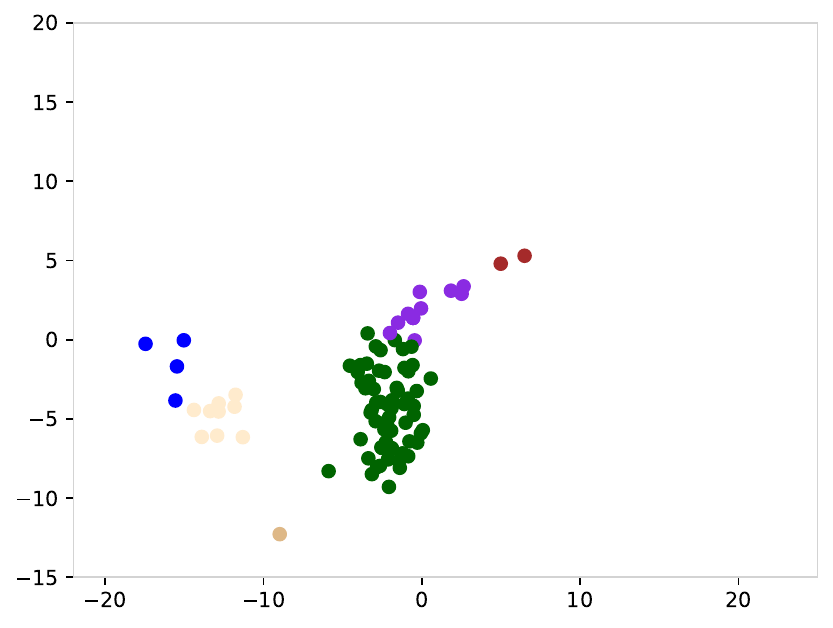}
	}
	\subfloat[Mixed adversary mode]{
		\includegraphics[width=0.33\linewidth]{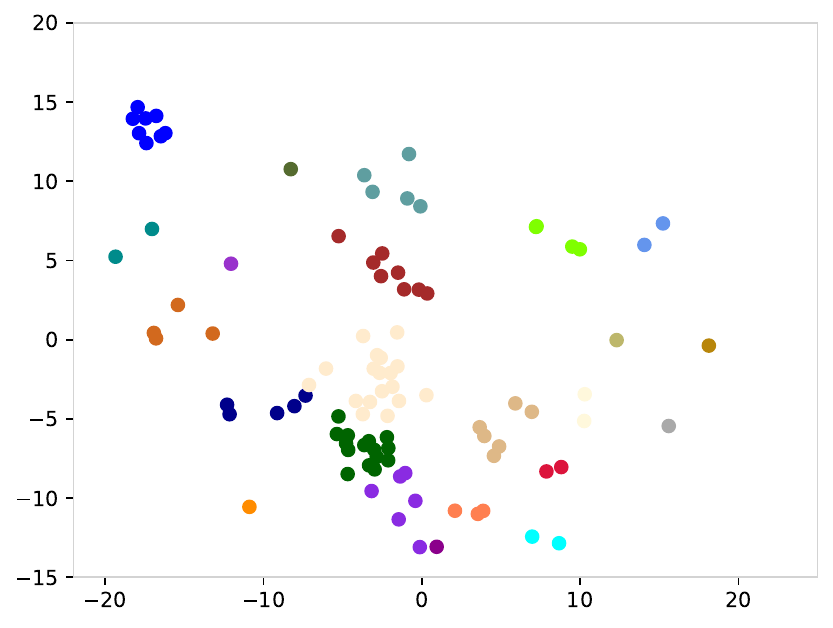}
	}
	\subfloat[Paired adversary mode]{
		\includegraphics[width=0.33\linewidth]{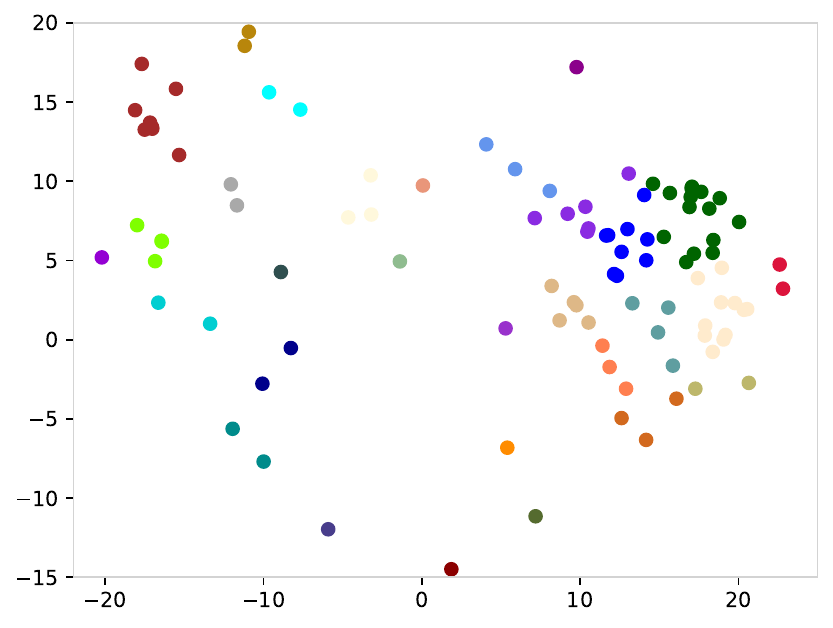}
	}
	\caption{The visualization of joint action distributions in three modes. Different clusters are with different colors. Please see details in Section~\ref{sec:adversary_competing}. }
	\label{action_list_comparison}
\end{figure*}

As discussed above, our developing SC2BA platform can be used as a benchmark to evaluate/foster new MARL algorithms. At the same time, one argued question is whether the models trained in the algorithm adversary mode have a better generalization, e.g., adapting to an unknown opponent or built-in AI bot. In this section, we attempt to answer this question from two aspects: increasing opponent diversity and enhancing adversary learning. The experiment results are reported in Fig.~\ref{improve_opponent_diverse}-Fig.~\ref{action_list_comparison}, whose itemized details are deferred to the supplementary file.

\subsubsection{Opponents Diversity}
\label{sec:opponents diversity}

We conduct experiments to validate whether increasing opponent diversity can enhance the algorithm performance adversarial to built-in AI bot. We use the algorithms, DOP and QMIX, as the red team respectively, where DOP and QMIX are the most excellent policy-based and value-based algorithms in SC2BA. There are two training modes: built-in AI bots and multi-algorithm mixed adversary. In the first mode, the algorithm are trained by engaging in battles against built-in AI bots, same to the SMAC setting. In the second mode, the opponents consist of a combination of fixed well-trained models as used in Section~\ref{sec:mixed_adversary_mode}. The test is to compete with built-in AI bots, i.e., validating the generalization ability to built-in AI bots. The results are illustrated in Fig.~\ref{improve_opponent_diverse}, from which we can observe increasing opponent diversity can boost the performance in most cases. In the 8m, 5m\_vs\_6m, and 10m\_vs\_11m scenarios, the mixed adversary mode facilitates faster discovery of winning strategies than built-in AI mode.  For the 8m scenario, the model trained in built-in AI mode achieves 100\% win rates, and the model trained by the mixed adversary mode has a little degradation at the terminate step, which might be attributed to the diversity of training opponents and the singularity of test opponent. 
These results indicate that increasing opponent diversity can indeed boost MARL algorithm performances.

\subsubsection{Adversary Competing}
\label{sec:adversary_competing}
In this section, our purpose is to validate what training mode is more effective
for MARL model. we test the adversary capability of one model (w.r.t one algorithm) when trained by three different modes: dual-algorithm adversary, mixed algorithm adversary and built-in AI bots, where the former two modes come from our SC2BA and the last one is from SMAC. Taking \{QMix, DOP\} as an example, we first train three models based on these three modes, denoted as three groups: \{QMix\_{dual}, DOP\_{dual}\}, \{QMix\_{mixed}, DOP\_{mixed}\} and \{QMix\_{AI}, DOP\_{AI}\}, then let every model of the former two groups compete with everyone of the last one group, such as QMix\_{dual} vs QMix\_{AI}, {QMix\_{dual} vs DOP\_{AI}, etc. We accumulate all experimental results of eight MRAL algorithms to report in Fig.~\ref{multi_paired_generalize} (each itemized results are given in the supplemental file). The blue box and the green box respectively represent the challenger results of agents training by two modes of SC2BA, i.e., dual-algorithm and mixed-algorithm adversaries. 
We observe that the agents trained by paired and mixed adversary mode have a higher win rate than the challenger that trained only built-in AI bots.
Further, the agents trained in paired mode perform slightly better, compared to the mixed adversary mode. This can be attributed to the use of evolvable adversaries, i.e., dynamic opponent. As two competitive models need to adapt to each other, the struggle of strategy and counter-strategy implicitly increases the diversity/complexity of policy learning, which could improve the learned model to some extent. The experiments provide some evidence that the dual-algorithm adversary mode can boost the generalization capability, making agents possess strong adversarial ability even in the face of unseen opponents. 

To further study why agents trained in the two modes of SC2BA perform well, we exhibit the behaviors of agents trained in these two modes (vs built-in AI bots), as shown in Fig.~\ref{action_list_comparison}. Concretely, we collect a joint action list of multi-agent from 100 episodes and execute principle component analysis (PCA) for a 2D space followed by the MeanShift~\cite{cheng1995mean} clustering for convenient visual exhibition.
In contrast to built-in AI mode, the actions learnt in the mixed and paired modes have more diversity. And, the paired adversary mode has a little higher diversity than the mixed mode. All these phenomena match the above observation in Fig.~\ref{multi_paired_generalize}.
It indicates that the the adversarial modes in SC2BA can benefit the policy diversity and be used to improve the learning of multi-agent.
	
\section{Conclusion and Future Work}

In this work, we created the SC2BA environment and APyMARL library to benchmark MARL algorithms in an adversary paradigm. The APyMARL was developed as an open-source library with easy-to-use interfaces/configurations and flexible modules, so as to serve SC2BA. Built on the popular RTS game StarCraft II, SC2BA is equipped with two types of algorithm-adversary modes: dual-algorithm paired adversary mode and multi-algorithm mixed adversary mode. In the former mode, both teams of the battle may be manipulated with different MARL algorithms, not that one team is fixed as built-in AI bots as in SMAC. That is, one algorithm needs to be competed with another algorithm. Hence, their learnability as well as dynamicity would cause not only the difficulty of optimization but also the diversity of policy actions. Hereby, it would result into an interesting and diversified evaluation on MARL algorithms. In the latter mode, we attempted to endow enemy units with multiple opponents modeled by multiple algorithms, which further enriches the competitive policy space. Based on the two adversarial modes, we provided extensive benchmarking tests for those representative MARL algorithms as well as symmetric/asymmetric scenarios. In our experience, we found that the algorithm-adversary could increase the diversity of policy, improve the robustness of model, and enhance the generalization ability of model. 

Further, we found some thought-provoking problems in algorithm adversaries, such as the sensibility to scenario layout (e.g., troops), the difficulty of heterogeneous scenarios, the fluctuation of algorithm adversary, the joint evolution of multi-opponents, etc. These issues seem to have a few focuses or studies in the field of MARL, their solutions need not only new proposals of MARL algorithms but also rethink some internal rules (e.g., reward function, motion law of agents, situation law, etc) related to game mechanism.

In summary, this SC2BA platform would facilitate the evaluation of MARL algorithms from the new different view of algorithm-adversary, and meantime provide support for cultivating new MARL algorithms. In the future, we will attempt to explore dynamic multi-algorithm mixed adversary, and design more asymmetric scenarios to further optimize the SC2BA platform. 


\bibliographystyle{IEEEtran}
\bibliography{SMAB_reference.bib}


\end{document}